\documentclass[letterpaper, 10 pt, conference]{IEEEtran}

\IEEEoverridecommandlockouts

\usepackage{comment}
\usepackage[vlined,ruled,linesnumbered]{algorithm2e}
\usepackage{graphics} %
\usepackage{rotating}
\usepackage{color}
\usepackage{enumerate}
\usepackage[T1]{fontenc}
\usepackage{psfrag}
\usepackage{epsfig} %
 \usepackage{hyperref}
\usepackage{booktabs}
\usepackage{graphicx,url}
\usepackage{multirow}
\usepackage{array}
\usepackage{latexsym}
\usepackage{amsfonts}
\usepackage{amsmath}
\usepackage{amssymb}
\usepackage{amsthm}
\usepackage[noend]{algorithmic}
\usepackage{xcolor}
\usepackage{prettyref}
\usepackage{flexisym}
\usepackage{bigdelim}
\usepackage{breqn} %
\usepackage{listings}
\usepackage{enumitem}
\usepackage{xspace}
\usepackage{bm}
\graphicspath{{./figures/}}
\usepackage{tikz}
\usetikzlibrary{matrix,calc}

\usepackage{amsthm,amsmath}
\newrefformat{prob}{Problem\,\ref{#1}}
\newrefformat{def}{Definition\,\ref{#1}}
\newrefformat{sec}{Section\,\ref{#1}}
\newrefformat{sub}{Section\,\ref{#1}}
\newrefformat{prop}{Proposition\,\ref{#1}}
\newrefformat{app}{Appendix\,\ref{#1}}
\newrefformat{alg}{Algorithm\,\ref{#1}}
\newrefformat{cor}{Corollary\,\ref{#1}}
\newrefformat{thm}{Theorem\,\ref{#1}}
\newrefformat{lem}{Lemma\,\ref{#1}}
\newrefformat{fig}{Fig.\,\ref{#1}}
\newrefformat{tab}{Table\,\ref{#1}}

\newtheorem{theorem}{Theorem}
\newtheorem{problem}{Problem}

\newtheorem{proposition}[theorem]{Proposition}
\newtheorem{remark}[theorem]{Remark}

\newcommand{\cf}{\emph{cf.}\xspace}

\newcommand{\bdmath}{\begin{dmath}}
\newcommand{\edmath}{\end{dmath}}
\newcommand{\beq}{\begin{equation}}
\newcommand{\eeq}{\end{equation}}
\newcommand{\bdm}{\begin{displaymath}}
\newcommand{\edm}{\end{displaymath}}
\newcommand{\bea}{\begin{eqnarray}}
\newcommand{\eea}{\end{eqnarray}}
\newcommand{\beal}{\beq \begin{array}{ll}}
\newcommand{\eeal}{\end{array} \eeq}
\newcommand{\beas}{\begin{eqnarray*}}
\newcommand{\eeas}{\end{eqnarray*}}
\newcommand{\ba}{\begin{array}}
\newcommand{\ea}{\end{array}}
\newcommand{\bit}{\begin{itemize}}
\newcommand{\eit}{\end{itemize}}
\newcommand{\ben}{\begin{enumerate}}
\newcommand{\een}{\end{enumerate}}

\newcommand{\calL}{{\cal L}}
\newcommand{\calM}{{\cal M}}

\newcommand{\calR}{{\cal R}}

\newcommand{\setal}{~\emph{et~al.}\xspace}
\newcommand{\eg}{\emph{e.g.,}\xspace}
\newcommand{\ie}{\emph{i.e.,}\xspace}
\newcommand{\myParagraph}[1]{{\bf #1.}\xspace}

\newcommand{\M}[1]{{\bm #1}} 
\renewcommand{\boldsymbol}[1]{{\bm #1}}

\newcommand{\RT}[1]{{\color{blue} \textbf{RT}: #1}}

\newcommand{\hide}[1]{}

\newcommand{\hiddenText}{{\color{gray} hidden text.}}
\newcommand{\hideWithText}[1]{\hiddenText}

\newcommand{\kron}{\otimes}

\DeclareMathOperator*{\argmin}{arg\,min}

\newcommand{\tran}{^{\mathsf{T}}}

\newcommand{\inv}{^{-1}}

\newcommand{\ones}{{\mathbf 1}}

\newcommand{\Real}[1]{ { {\mathbb R}^{#1} } }

\newcommand{\SEthree}{\ensuremath{\mathrm{SE}(3)}\xspace}

\newcommand{\MB}{\M{B}}

\newcommand{\MM}{\M{M}}

\newcommand{\MR}{\M{R}}

\newcommand{\MI}{\M{I}}

\newcommand{\MO}{\M{O}}
\newcommand{\MT}{\M{T}}
\newcommand{\MX}{\M{X}}

\newcommand{\vb}{\boldsymbol{b}}

\newcommand{\vf}{\boldsymbol{f}}

\newcommand{\vn}{\boldsymbol{n}}

\newcommand{\vu}{\boldsymbol{u}}

\newcommand{\vt}{\boldsymbol{t}}
\newcommand{\vxx}{\boldsymbol{x}} 
\newcommand{\vy}{\boldsymbol{y}}

\newcommand{\blue}[1]{{\color{blue}#1}}

\newcommand{\linkToPdf}[1]{\href{#1}{\blue{(pdf)}}}
\newcommand{\linkToPpt}[1]{\href{#1}{\blue{(ppt)}}}
\newcommand{\linkToCode}[1]{\href{#1}{\blue{(code)}}}
\newcommand{\linkToWeb}[1]{\href{#1}{\blue{(web)}}}
\newcommand{\linkToVideo}[1]{\href{#1}{\blue{(video)}}}
\newcommand{\linkToMedia}[1]{\href{#1}{\blue{(media)}}}
\newcommand{\award}[1]{\xspace} 

\usepackage{wrapfig}
\usepackage{caption}
\usepackage{subcaption}
\usepackage[capitalise]{cleveref}
\usepackage{xr}

\makeatletter
\newcommand{\printfnsymbol}[1]{%
	\textsuperscript{\@fnsymbol{#1}}%
}
\makeatother

\newcommand{\finalisarxivbutomit}[1]{{#1}}
\newcommand{\arxivadd}[1]{{#1}}
\newcommand{\arxivomit}[1]{{}}

\arxivomit{\externaldocument{supplementary}}

\title{A Correct-and-Certify Approach to Self-Supervise Object Pose Estimators via Ensemble Self-Training}

\author{Jingnan Shi\printfnsymbol{1}\thanks{\printfnsymbol{1} equal contribution}, Rajat Talak\printfnsymbol{1}, Dominic Maggio, and Luca Carlone
\thanks{This work was partially funded by Army Research Laboratory
		Distributed and Collaborative Intelligent Systems and Technology Collaborative
		Research Alliance under Grant W911NF-17-2-0181, in part by the Office of Naval Research RAIDER project under Grant N00014-18-1-2828, in part by the NSF CAREER award “Certifiable Perception for Autonomous Cyber-Physical Systems,” and the Amazon project “Next-Generation Spatial AI for Human-Centric Robotics.”}
\thanks{J.\,Shi, R.\,Talak, D.\ Maggio, L.\,Carlone are with the Laboratory for
	Information \& Decision Systems (LIDS), Massachusetts Institute of Technology, Cambridge, MA 02139, USA, Email: {\texttt \{jnshi,talak,drmaggio,lcarlone\}@mit.edu}}

}

\begin{document}

\maketitle
\finalisarxivbutomit{
\arxivadd{
\begin{tikzpicture}[overlay, remember picture]
\path (current page.north east) ++(-3.2,-0.4) node[below left] {
This paper has been accepted for publication at the 2023 Robotics: Science and Systems Conference.
};
\end{tikzpicture}
\begin{tikzpicture}[overlay, remember picture]
\path (current page.north east) ++(-2,-0.8) node[below left] {
Please cite the paper as: J. Shi*, R. Talak*, D. Maggio and L. Carlone, ``A Correct-and-Certify Approach to Self-Supervise
};
\end{tikzpicture}
\begin{tikzpicture}[overlay, remember picture]
\path (current page.north east) ++(-3.5,-1.2) node[below left] {
Object Pose Estimators via Ensemble Self-Training,'' \emph{Robotics: Science and Systems (RSS)}, 2023.
};
\end{tikzpicture}
}
}

\newcommand{\name}{{Ensemble}\xspace}
\newcommand{\nameCsy}{Ensemble-\csy}
\newcommand{\nameRky}{Ensemble-\rKeyPo}
\newcommand{\cpo}{C3PO\xspace}
\newcommand{\cpopp}{C3PO++\xspace}
\newcommand{\KeyPo}{KeyPo\xspace}
\newcommand{\rKeyPo}{RKN\xspace}
\newcommand{\rKeyPoLong}{Robust Keypoint Network\xspace}
\newcommand{\csy}{CosyPose\xspace}

\newcommand{\robustCorrector}{Robust Corrector\xspace}
\newcommand{\naive}{Naive\xspace}
\newcommand{\naiveICP}{Naive + ICP\xspace}

\newcommand{\ES}{Ensemble Self-Training\xspace}
\newcommand{\Es}{Ensemble self-training\xspace}
\newcommand{\es}{ensemble self-training\xspace}
\newcommand{\ensemble}{ensemble\xspace}
\newcommand{\Ensemble}{Ensemble\xspace}

\newcommand{\cTwoD}{2D}
\newcommand{\cThreeD}{3D}

\newcommand{\YCBV}{YCBV\xspace}
\newcommand{\TLESS}{TLESS\xspace}

\newcommand{\selfSD}{Self6D++\xspace}
\newcommand{\corrected}{corrector\xspace}

\newcommand{\outperforms}{performs on par or better than\xspace}

\newcommand{\inputPC}{\ensuremath{\MX}\xspace}
\newcommand{\inputRGBD}{\ensuremath{\MI}\xspace}
\newcommand{\inputMask}{\ensuremath{\MM}\xspace}
\newcommand{\inputFeat}{\ensuremath{\vf}\xspace}
\newcommand{\inpuT}{\ensuremath{(\inputRGBD, \inputMask, \inputPC, \inputFeat)}\xspace}

\newcommand{\gtR}{\ensuremath{\MR^{\ast}}\xspace}
\newcommand{\gtt}{\ensuremath{\vt^{\ast}}\xspace}
\newcommand{\gtKP}{\ensuremath{\vy^{\ast}}\xspace}
\newcommand{\gtPose}{\ensuremath{\MT^{\ast}}\xspace}
\newcommand{\gtMask}{\ensuremath{\MM^{\ast}}\xspace}

\newcommand{\noiseMask}{\ensuremath{\MO_{\MM}}\xspace}
\newcommand{\noisePC}{\ensuremath{\vn_w}\xspace}
\newcommand{\outliersPC}{\ensuremath{\MO_{\MX}}\xspace}
\newcommand{\cameraMap}[1]{\ensuremath{\pi\inv\left(#1\right)}\xspace}

\newcommand{\render}[1]{\ensuremath{\calR}\left(#1\right)}

\newcommand{\occFun}[1]{\ensuremath{\Theta\left(#1\right)}\xspace}

\newcommand{\nrKeypoints}{\ensuremath{N}\xspace}
\newcommand{\nrPoints}{\ensuremath{n}\xspace}
\newcommand{\kp}{\ensuremath{\vb}\xspace}
\newcommand{\cad}{\ensuremath{\MB}\xspace}
\newcommand{\cadPC}{\ensuremath{\hat{\MB}}\xspace}

\newcommand{\estR}{\ensuremath{\hat{\MR}}\xspace}
\newcommand{\estt}{\ensuremath{\hat{\vt}}\xspace}
\newcommand{\detPose}{\ensuremath{\tilde{\MT}}\xspace}
\newcommand{\estPose}{\ensuremath{\hat{\MT}}\xspace}
\newcommand{\estMask}{\ensuremath{\hat{\MM}}\xspace}
\newcommand{\estImage}{\ensuremath{\hat{\MI}}\xspace}

\newcommand{\kpCorrection}{\ensuremath{\Delta \vy}\xspace}
\newcommand{\kpOptCorrection}{\ensuremath{\Delta \vy^{\ast}}\xspace}
\newcommand{\kpCorrected}{\ensuremath{\hat{\vy}}\xspace}

\newcommand{\detKP}{\ensuremath{\tilde{\vy}}\xspace}
\newcommand{\predKP}{\ensuremath{\hat{\vy}}\xspace}
\newcommand{\predCAD}{\ensuremath{\hat{\MX}}\xspace}

\newcommand{\zetaCorrectness}{$\zeta$-correctness\xspace}
\newcommand{\zetaCorrect}{$\zeta$-correct\xspace}

\newcommand{\ocThreeD}[2]{\ensuremath{\texttt{oc}_{\text{3D}}(#1, #2)}\xspace}
\newcommand{\ocTwoD}[2]{\ensuremath{\texttt{oc}_{\text{2D}}(#1, #2)}\xspace}
\newcommand{\oc}[2]{\ensuremath{\texttt{oc}(#1, #2)}\xspace}
\newcommand{\ocx}{\ensuremath{\texttt{oc}}\xspace}
\newcommand{\ocThreeDx}{\ensuremath{\texttt{oc}_{\text{3D}}}\xspace}
\newcommand{\ocTwoDx}{\ensuremath{\texttt{oc}_{\text{2D}}}\xspace}

\newcommand{\Mone}{\ensuremath{\calM^1}\xspace}
\newcommand{\Mtwo}{\ensuremath{\calM^2}\xspace}
\newcommand{\detKPone}{\ensuremath{\detKP^1}\xspace}
\newcommand{\detKPtwo}{\ensuremath{\detKP^2}\xspace}
\newcommand{\detPoseOne}{\ensuremath{\detPose^1}\xspace}
\newcommand{\detPoseTwo}{\ensuremath{\detPose^2}\xspace}
\newcommand{\predKPone}{\ensuremath{\predKP^1}\xspace}
\newcommand{\predKPtwo}{\ensuremath{\predKP^2}\xspace}
\newcommand{\estPoseOne}{\ensuremath{\estPose^1}\xspace}
\newcommand{\estPoseTwo}{\ensuremath{\estPose^2}\xspace}
\newcommand{\outOne}{\ensuremath{(\detKPone, \detPoseOne)}\xspace}
\newcommand{\outTwo}{\ensuremath{(\detKPtwo, \detPoseTwo)}\xspace}
\newcommand{\correctedOne}{\ensuremath{(\predKPone, \estPoseOne)}\xspace}
\newcommand{\correctedTwo}{\ensuremath{(\predKPtwo, \estPoseTwo)}\xspace}
\newcommand{\correctedOutput}{\ensuremath{(\predKP, \estPose)}\xspace}

\newcommand{\scoreSet}[2]{\ensuremath{\mathbb{S}(#1, #2)}\xspace}
\newcommand{\distHausdorff}[2]{\ensuremath{d_{H}(#1, #2)}}
\newcommand{\indicator}[1]{\ensuremath{\mathbb{I}\left\{ #1 \right\}}}
\newcommand{\mlp}[1]{\ensuremath{\text{MLP}\left(#1\right)}\xspace}
\newcommand{\mlpx}{\ensuremath{\text{MLP}}\xspace}
\newcommand{\area}[1]{\ensuremath{\text{ar}\left( #1 \right)}\xspace}

\newcommand{\JS}[1]{#1}

\begin{abstract}
Real-world robotics applications demand object pose estimation methods that work reliably across a variety of scenarios. Modern learning-based approaches require large labeled datasets and tend to perform poorly outside the training domain.
Our first contribution is to develop a \emph{robust corrector} module that corrects pose estimates using depth information, thus enabling existing methods to better generalize to new test domains; 
 the corrector operates on semantic keypoints (but is also applicable to other pose estimators) and is fully differentiable.
 Our second contribution is an \emph{\es} approach that simultaneously trains multiple pose estimators in a self-supervised manner. Our \es architecture uses the robust corrector to refine the output of each pose estimator; then, it evaluates the quality of the outputs using \emph{observable correctness} certificates;
  finally, it uses the observably correct outputs for further training, without requiring external supervision. 
As an additional contribution, we propose small improvements to a regression-based 
keypoint detection architecture, to enhance its robustness to outliers; these improvements include a 
robust pooling scheme and a robust centroid computation.
Experiments on the \YCBV and \TLESS datasets show 
the proposed \es \outperforms fully supervised baselines while not requiring 3D annotations on real data.
Code and pre-trained models are available on the project web page\footnote{\url{https://web.mit.edu/sparklab/research/ensemble_self_training/}}.
\end{abstract}

%
%
%
%

 %
%

\section{Introduction}

 Object pose estimation is a key capability for robotics.
From search-and-rescue in caves and subways~\cite{Ebadi22arxiv-surveySLAMSubt}, to domestic robotics and factory automation~\cite{Zeng17icra-amazonChallenge}, 
to satellite pose estimation for autonomous docking and debris removal~\cite{Chen19ICCVW-satellitePoseEstimation}, estimating 3D poses of external objects is a necessary prerequisite for robot navigation, manipulation, and human-robot interaction.

Recent progress in learning-based object pose estimation has been fueled by the availability of pose-annotated datasets~\cite{Calli15ram-BenchmarkingManipulation, Hodan17-TLESSRGBD} and computer vision challenges 
(\eg the BOP challenge~\cite{Hodan18eccv-BOPBenchmark, Hodan20eccvw-BOPChallenge}).    
However, in robotics applications it is not always practical to collect training data for all relevant environments, also considering that most techniques require \emph{3D} pose annotations, which are more challenging to manually annotate. 
While increasingly more realistic simulators are reducing the sim-to-real gap, 
  it remains unclear if the gap will eventually disappear, and while simulation makes it easy to 
  provide ground-truth labels for the object poses, creating a variety of assets to cover 
  different scenarios remains an expensive and time-consuming task.
Therefore, it would be desirable to develop self-supervised learning approaches that can 
learn from real data without manual 3D annotations.

A related issue is that current learning-based techniques generalize poorly outside the training domain. 
For instance, a change in the placement of the camera (\eg from a ground robot to a drone) may induce a domain gap. Similarly, an approach trained in simulation might exhibit a sim-to-real gap, hindering real-world performance.
Therefore, it would be desirable to design approaches that can better bridge the gap between training and testing, and generalize to unseen data.

\begin{figure}[t]
\vspace{-5mm}
	\centering
	\includegraphics[trim=0 0 0 0,clip,width=0.9\linewidth]{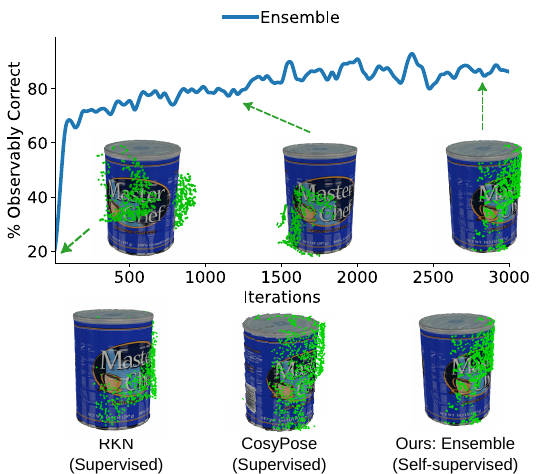}
	\caption{
    We propose an \emph{\es} architecture that simultaneously trains multiple pose estimators without external supervision. (\textit{Top}) The percentage of observably correct pose estimates increases as the self-training progresses, jumping from $19\%$ to $80\%$ after $3000$ iterations.
    (\textit{Bottom}) The resulting estimates are on par or better compared to fully supervised baselines,
    including a point-cloud based architecture (\rKeyPo) and \csy~\cite{Labbe20eccv-CosyPose}. \JS{Green points represent the input point cloud transformed into the model frame based on the estimated pose.}
    \label{fig:front-page} \vspace{-8mm}}
\end{figure}

In this paper we tackle these issues with two main contributions. 
First, we propose a {\bf robust corrector} (\cref{sec:robustCorrector}), which can be added to the output of any pose estimation architecture 
(as long as a CAD model of the object is available) and is able to correct moderate pose estimation errors. The robust corrector  is a differentiable optimization layer, and corrects the pose output by using depth or point-cloud data and the given object CAD model. While Talak\setal~\cite{Talak23arxiv-c3po} showed that a similar scheme outperforms standard pose refinement methods (\eg ICP), the key feature of our method is its robustness to outliers, \ie points incorrectly segmented to be part of the object. Additionally, the robust corrector enables a simple differentiation rule for back-propagation. %

Our second contribution is an {\bf \es} approach (\cref{sec:parallel}), in which, any number of existing object pose estimation models can be augmented and self-trained, in parallel, on real-world data without requiring pose annotations. 
The \es approach attaches a robust corrector to each pose estimator in order to refine their outputs.
Then, it computes \emph{certificates of observable correctness} to assess the quality of the output of each pose estimator. An observable correctness certificate is a binary condition that checks whether a pose estimate (after being refined by the corrector) meets some basic geometric consistency requirements. We impose two such consistency checks. The first, requires that the posed CAD model (posed at the \corrected pose estimate) matches the input. The second, requires that a rendering of the posed CAD model matches a 2D segmentation mask of the object. These certificates extend the corresponding observable correctness certificates in~\cite{Talak23arxiv-c3po} to account for the presence of outliers and to operate on both 3D and 2D data.
The \corrected pose, output by the robust corrector, is said to be observably correct, if it meets both these geometric consistency checks. 
Finally, the \es approach selects the observably correct outputs across all the pose estimators in the \ensemble, and uses them for further self-supervised training. The key idea is that when an estimator produces an observably correct output, it can learn from it, while simultaneously informing the other estimators. This allows exploiting the complementarity between different models (\eg a point-cloud-based architecture might perform better than an RGB-based one in low-lighting conditions), while helping to
bootstrap the self-training process (\eg if a model initially is not able to get observably correct instances, 
 it still gets informed by observably correct instances produced by the other models).  
 In our experiments, we observe that as the self-training progresses the number of observably correct instances increases, and the proposed \es approach outperforms each model in the \ensemble, even when those are provided with full supervision (\cref{fig:front-page}).

As an addition contribution, we develop an {\bf outlier-robust point cloud processing} by proposing 
 small improvements to a standard point-cloud-based regression  model for semantic-keypoint detection (\cref{sec:rkeypo}).
Regression models for semantic-keypoint detection (\eg~\cite{Talak23arxiv-c3po}) are not robust to outliers in the segmentation mask due to two reasons:
(i) most of the point cloud architectures use farthest point sampling (FPS)~\cite{Qi17cvpr-pointnet, Qi17nips-pointnet++, Zhao21iccv-PointTransformer}, which in the presence of outliers ends up sampling many outlier points, rather than rejecting them; (ii) the lack of exact translation and rotation equivariance of point-cloud-based architectures results in incorrect detections, when the input point cloud is not centered correctly --- a phenomenon that happens often when the point cloud contains many outliers.
We propose a trainable {\bf robust pooling} layer and a {\bf robust centroid} computation, that can make any point cloud-based model robust to outliers. The robust pooling samples points based on regressed scores, using point features and trainable weights. The robust centroid computation computes a robust center for the input point cloud using graduated non-convexity~\cite{Yang20ral-GNC}. The point cloud is then centered at this robust center, before passing it to the point cloud-based model.

We conclude the paper by evaluating the robust corrector, the \es, and the outlier-robust point cloud processing on the YCBV~\cite{Calli15-YCBobject, Hodan20eccvw-BOPChallenge} and the TLESS~\cite{Hodan17-TLESSRGBD, Hodan20eccvw-BOPChallenge} datasets. 
The results confirm that (i)~the robust corrector is able to correct large errors in the keypoint detections; (ii)~the \es produces pose estimates on par or better compared to fully supervised baselines (10\% average increase in the ADD-S (AUC) scores over CosyPose~\cite{Labbe20eccv-CosyPose} on selected \YCBV objects), and largely outperforms competing self-supervised methods (40\% increase in the ADD-S (AUC) scores over~\selfSD~\cite{Wang22pami-OcclusionAwareSelfSupervised}); (iii)~the outlier-robust point cloud processing increases the robustness of the keypoint detection in the presence of outliers.
The code will be released upon the acceptance of this paper. 

Before delving into our contributions, we review related work (\cref{sec:relatedWork}) and formally state the problem (\cref{sec:prob}).

\section{Related Works}
\label{sec:relatedWork}

\myParagraph{Self-Supervised Object Pose Estimation}
The literature on self-supervised pose estimation is sparse and very recent.
Wang\setal~\cite{Wang20eccv-Self6DSelfSupervised} train a pose estimation model on synthetic RGB-D data, and then refine it further with self-supervised training on real, unannotated data; differentiable rendering provides the required supervision signal. 
\selfSD~\cite{Wang22pami-OcclusionAwareSelfSupervised}  extends~\cite{Wang20eccv-Self6DSelfSupervised} 
by accounting for possible occlusions of the object.
Chen\setal~\cite{Chen22eccv-SimtoReal6D} propose a student-teacher iterative scheme to bridge the sim-to-real domain gap, for a sim-trained pose estimation model. %
Wang\setal~\cite{Wang19-normalizedCoordinate} extract pose-invariant features, thereby canonizing object's shape, and use them for category-level object pose estimation of unseen object instances.
Zakharov\setal~\cite{Zakharov20cvpr-Autolabeling3D} utilize differentiable rendering of signed distance fields of objects, along with normalized object coordinate spaces~\cite{Wang19-normalizedCoordinate}, to learn 9D cuboids in a self-supervised manner.
Zhang\setal~\cite{Zhang23iclr-SelfsupervisedGeometric} propose to jointly reconstruct the 3D shape of an object category and learn dense 2D-3D correspondences between the input image and the 3D shape, by training on large-scale, real-world object videos.
Deng\setal~\cite{Deng20icra-Selfsupervised6D} self-supervise pose estimation by interacting with the objects in the environment; the model gets trained on the data collected autonomously by 
a manipulator.

Point-cloud-only, self-supervised approaches for pose estimation  have been also proposed.
Li\setal~\cite{Li21nips-LeveragingSE} extract an \SEthree-invariant feature, which works as a canonical object, and use it to supervise training with a Chamfer loss.  Sun\setal~\cite{Sun21nips-CanonicalCapsules} tackle self-supervised point cloud alignment by extracting features using capsule network. Talak\setal~\cite{Talak23arxiv-c3po} propose a self-supervised keypoint-based pose estimator that uses a corrector and binary certificates during self-supervision.

Our approach shares insights with~\cite{Chen22eccv-SimtoReal6D,Wang20eccv-Self6DSelfSupervised,Wang22pami-OcclusionAwareSelfSupervised,Talak23arxiv-c3po}. Similar to these approaches, our goal is to take sim-trained models and further train them in a self-supervised manner. 
The works~\cite{Wang22pami-OcclusionAwareSelfSupervised,Chen22eccv-SimtoReal6D} use a student-teacher architecture which requires instantiating two networks: the teacher and the student; on the other hand, with the proposed corrector and certificates we can directly self-supervise the pose estimation models without the need to instantiate another network.
The works~\cite{Talak23arxiv-c3po,Chen22eccv-SimtoReal6D} use certificates or consistency checks to assess the quality of the pose estimates for self-training; contrary to our certificates, the ones in~\cite{Talak23arxiv-c3po,Chen22eccv-SimtoReal6D} are sensitive to outliers resulting from a noisy segmentation of the object, due to their reliance on non-robust distance functions. 
Conversely,~\cite{Wang20eccv-Self6DSelfSupervised} is not robust to foreground occlusions of the object, as shown in~\cite{Wang22pami-OcclusionAwareSelfSupervised}.
 The work~\cite{Talak23arxiv-c3po} is the first to propose a self-training procedure based on a corrector and binary certificates. However, the approach is restricted to keypoint-based methods and performs poorly in the presence of outliers (\cref{sec:expt}).
 In addition, none of the works above jointly self-supervise multiple models. %

\myParagraph{Point Cloud Architectures} The success of convolutional neural networks on images has led many researchers to investigate models that can work directly on point cloud data (\eg produced by a LiDAR or an RGB-D camera). However, most existing point-cloud-based models suffer from the fact that there is not an 
easy and efficient equivalent to the pooling/unpooling operation for point cloud data.
Most models use farthest point sampling (FPS) (which sequentially samples a set of points, such that each point is farthest away from all the points sampled thus far). 
This method often leads to 
sub-optimal performance~\cite{Wang22arxiv-LighTNLightweight, Lin22tcvg-TaskAwareSampling, Qian21arxiv-MOPSNetMatrix, Yan20cvpr-PointASNLRobust, Nezhadarya20cvpr-AdaptiveHierarchical, Lang20cvpr-SampleNetDifferentiable, Yang19cvpr-ModelingPoint, Dovrat19cvpr-LearningSample}, especially in the presence of outliers; see also our analysis in~\cref{sec:expt}. 
Dovrat\setal~\cite{Dovrat19cvpr-LearningSample} show that a task-specific, learning-based approach to sub-sample points can outperform a method like FPS.
Yang\setal~\cite{Yang19cvpr-ModelingPoint} propose a learnable, task-agnostic, Gumbel subset sampling, which produces ``soft'' subsets in training, and hard discrete subsets at test-time.
Lang\setal~\cite{Lang20cvpr-SampleNetDifferentiable} propose a differentiable relaxation of point cloud sampling, as a mixture of points.
Nezhadarya\setal~\cite{Nezhadarya20cvpr-AdaptiveHierarchical} propose a global down-sampling method that first determines critical points to sample, from regressed point features, hence obviating the need to use computationally expensive k-nearest neighbor-based aggregation.
Yan\setal~\cite{Yan20cvpr-PointASNLRobust} propose adaptive sampling, that modifies the sampled points using FPS, and then uses a local-nonlocal module to extract and capture neighboring and long-range dependencies of the sampled points. 
Lin\setal~\cite{Lin22tcvg-TaskAwareSampling}  propose a learning-based sampling strategy, and shows how the sampled points differ across various tasks such as object-part segmentation and point cloud completion --- making the point that the sampling strategy needs to be task specific. 
Wang\setal~\cite{Wang22arxiv-LighTNLightweight} propose a lightweight transformer network for point cloud downsampling. 
None of these works 
investigate the effect of outliers in point-cloud-based regression models for semantic keypoint detection.%

\myParagraph{Differentiable Optimization}
A differentiable optimization layer solves an optimization problem, and also computes the gradient of the optimal solution with respect to the input parameters, for back-propagation.  
A differentiable optimization layer allows a learning-based model to explicitly take into account various geometric and physical constraints.
Related work has developed tools to differentiate through 
quadratic optimization problems~\cite{Amos17ICML-optnet},
convex problems~\cite{Agrawal19NIPS-diffConvex}, non-linear optimization problems~\cite{Gould22pami-DeepDeclarative}, stochastic optimization problems~\cite{Donti17nips-TaskbasedEndtoend}, 
and combinatorial optimization problems~\cite{Pogancic20iclr-DifferentiationBlackbox, Paulus21icml-CombOptNetFit,Wang19icml-SATNetBridging}.
Open-source libraries, such as~\cite{Wang22arxiv-PyPose, Pineda22nips-TheseusLibrary},  provide off-the-shelf tools to implement differentiable non-linear least-squares.
Our architecture implements a robust corrector module as a differentiable optimization layer. Although, it solves a non-convex optimization problem, we are able to exploit its specific structure to implement a very simple derivative.

 \begin{figure}[t]
    \centering
    \includegraphics[trim=0 0 0 0,clip,width=0.9\linewidth]{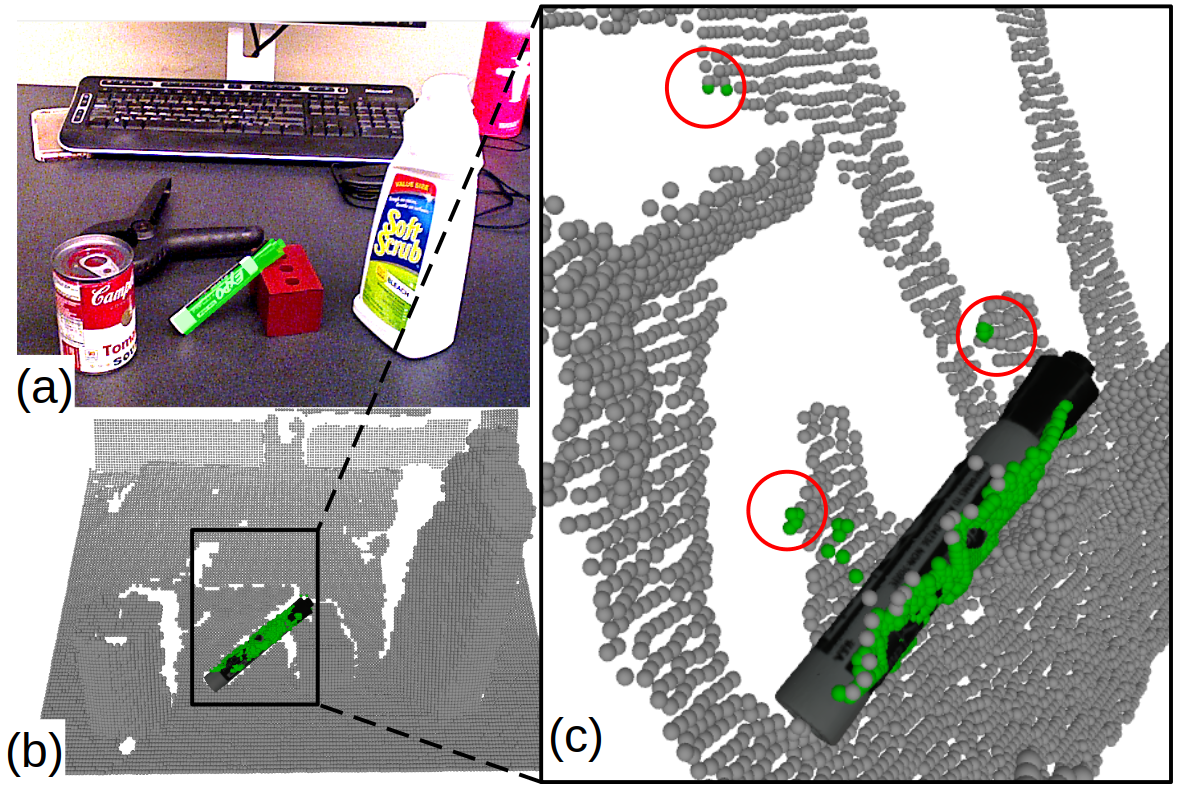}
    \caption{Examples of outliers caused by a noisy 2D segmentation. (a) RGB image with segmentation mask overlaid in green.
      (b) Depth map with masked points highlighted in green. (c) Zoomed-in view of the depth point cloud,
      with outliers circled in red.
      Outliers are common even in manually annotated masks: the example in the figure is obtained 
      using the ground-truth segmentation available in the \YCBV dataset.
    \vspace{-6mm}\label{fig:pc-outliers}}
\end{figure}

\section{Problem Statement}
\label{sec:prob}

Consider a robot equipped with a calibrated RGB-D camera. 
The camera collects color and depth data picturing a 3D scene containing an object of interest; we assume the object has a known shape, \ie we have its CAD model.
We also assume access to a standard pre-trained 2D object detection and instance segmentation model (\eg pre-trained MaskRCNN~\cite{He17iccv-maskRCNN}), and use it to extract 
a region of interest \inputRGBD  (RoI) and the object 
 2D segmentation mask \inputMask. Given the 2D segmentation mask and camera intrinsics, we compute the partial point cloud \inputPC of the detected object, along with its (RGB) color \inputFeat as:   
\begin{equation}
\label{eq:pc}
\inputPC, \inputFeat = \cameraMap{\inputRGBD, \inputMask},
\end{equation}
where \cameraMap{\cdot} simply back-projects the pixels in the mask \inputMask to 3D points using the depth and camera intrinsics, and attaches (RGB) color information to each point. Here, $\inputPC \in \Real{3 \times \nrPoints}$ and $\inputFeat \in 
\Real{3 \times \nrPoints}$, where \nrPoints is the number of points.

The segmentation mask \inputMask ---produced by the 2D segmentation model--- is typically noisy and may include outlier pixels that do not actually belong to the object or miss pixels that do belong to it (\cref{fig:pc-outliers}).
 More formally, if \inputMask is a segmented 2D binary mask, then we can write it as:
\begin{equation}
\inputMask = \gtMask + \noiseMask,
\end{equation}
where \gtMask denotes the ground-truth mask for the object and \noiseMask contains the misclassified (outlier) pixels.

The outliers in the segmentation mask, in turn, induce outliers in the partial point cloud \inputPC of the object, according to eq.~\eqref{eq:pc};
these are 3D points that do not actually belong to the object (\cref{fig:pc-outliers}). To formalize this concept, it is useful to define a generative model that relates the point cloud \inputPC with the to-be-computed object pose $\gtPose$ and its CAD model $\cad$. We write the generative model for the partial point cloud \inputPC as follows:
\begin{equation}
\label{eq:inputPCgen}
\inputPC = \occFun{\gtPose \cdot \cad} + \noisePC + \outliersPC,
\end{equation}
where $\gtPose \cdot \cad$ is the \emph{ground-truth posed CAD model} (\ie the CAD model arranged according to its true pose $\gtPose$),
\occFun{\cdot} is an occlusion function that samples a subset of points on the posed CAD model $\gtPose \cdot \cad$ of the object, restricted to the ground-truth mask \gtMask, \noisePC is the sensor measurement noise, and \outliersPC are the outlier points caused by outlier pixels in $\noiseMask$.

Our goal is to develop an estimator for the object pose $\gtPose$.  
We consider the realistic case where we have access to existing learning-based pose estimation models, but these have been trained on a different domain (\eg they have been trained in simulation). Therefore, our goal is to further train these models on real data without 3D annotations and in the presence of a potentially large domain gap (\eg sim-to-real gap). %
\begin{problem}[Self-Train Pose Estimation Models]
Propose a method to take one or more sim-trained 
pose estimation models, 
and train them, in parallel, with self-supervision  
on unannotated real-world data. 
\end{problem}
\section{Robust Corrector}
\label{sec:robustCorrector}

In this section we focus on keypoint-based pose estimation and we develop a \emph{robust corrector} that can compensate errors in the keypoint detections (\eg caused by the sim-to-real gap). In the next section we show that the same ideas can be applied to other pose estimators that do not rely on keypoints.

\begin{wrapfigure}{r}{0.32\linewidth}
	\centering 
	\vspace{1.5mm}
	\includegraphics[trim=80 85 100 50,clip,width=0.8\linewidth]{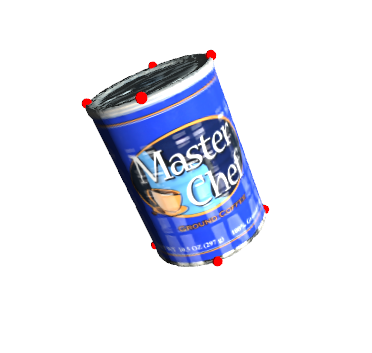}
	\caption{CAD model with annotated semantic keypoints.\label{fig:cad-kp}\vspace{-2mm}}
\end{wrapfigure}

 A semantic-keypoint-based pose estimator (\eg~\cite{Shi21rss-pace,Pavlakos17icra-semanticKeypoints,Schmeckpeper22arxiv-singleRGBpose}) 
 first uses a neural network to detect semantic keypoints \detKP from the input; 
 the neural network is trained to detect specific points on the surface of the object (see~\cref{fig:cad-kp}). 
 The detector can be implemented using CNN-based architectures on RGB inputs (\eg~\cite{Pavlakos17icra-semanticKeypoints,Schmeckpeper22arxiv-singleRGBpose}) or regression models (\eg point transformers~\cite{Zhao21iccv-PointTransformer}) on point cloud and RGB-D data. 
After detecting the keypoints, a pose estimate can be retrieved via point-cloud registration, \ie by computing the rigid transformation $\detPose$ that aligns the 
annotated keypoints $\kp$ on the CAD model (\cref{fig:cad-kp}), and the detected keypoints $\detKP$. 
In this section we will assume the keypoint detector to be given, while in~\cref{sec:parallel,sec:rkeypo}  we will discuss how to self-train a detector and how to improve the robustness of existing architectures, respectively.
\myParagraph{Robust Corrector Overview} The detected keypoints $\detKP$ may be inaccurate when there is a large domain gap, which in turn leads to poor pose estimates $\detPose$.
Our first contribution is a \emph{robust corrector}, that corrects the detected keypoints 
 by utilizing the dense information in the input point cloud \inputPC and the object CAD model \cad. 
Contrary to related work~\cite{Talak23arxiv-c3po},
the robust corrector is designed to automatically reject outliers in the input point cloud \inputPC, caused by the noisy segmentation mask \inputMask. 
In a nutshell, our robust corrector takes in the detected keypoints $\detKP$, and produces a correction \kpOptCorrection by solving the \emph{robust corrector optimization problem} (details below).
The corrected keypoints then become:
\begin{equation}
\label{eq:corrected-kp}
\predKP = \detKP + \kpOptCorrection.
\end{equation}

After computing the corrected keypoints $\predKP$, we can compute a
\emph{corrected pose} $\estPose(\kpOptCorrection)$ via outlier-free registration: 
\begin{equation}
\label{eq:corrected-pose}
\estPose(\kpOptCorrection) = \textstyle \argmin_{\MT \in \SEthree}~\sum_{i=1}^{\nrKeypoints}~\lVert \predKP[i] - \MT \cdot \kp[i] \rVert^{2}_2.
\end{equation}
where $\nrKeypoints$ is the number of semantic keypoints. 

We now describe the robust corrector optimization problem, and its forward and backward pass: since we 
are going to insert this module in a trainable architecture in~\cref{sec:parallel}, we are particularly interested in making it differentiable.

\myParagraph{Robust Corrector Optimization Problem} The robust corrector computes a correction term \kpCorrection, to the detected keypoints \detKP, by attempting to match the input point cloud to the resulting posed CAD model $\predCAD(\kpCorrection)$, which is arranged according to the corrected pose $\estPose(\kpOptCorrection)$.
 The corrector solves a bi-level optimization problem given by: 
\begin{equation}
\begin{aligned}
& \underset{\kpCorrection \in \Real{3\times \nrKeypoints}}{\text{Minimize}} 
& & \frac{1}{\nrPoints} \sum_{i=1}^n \rho\left( \min_{j}~|| \inputPC[i] - \predCAD(\kpCorrection)[j] ||_{2}\right), \label{eq:corrector} \\
& \text{subject to} %
&& \predCAD(\kpCorrection) = \MT(\kpCorrection) \cdot \cadPC, \\
&&&\hspace{-1.5cm}~\MT(\kpCorrection) = \argmin_{\MT \in \SEthree}~\sum_{i=1}^{\nrKeypoints}~\lVert \detKP[i] + \kpCorrection[i] - \MT \cdot \kp [i] \rVert^{2}_{2}  
\end{aligned}
\end{equation} 
where $\inputPC[i]$ and $\predCAD(\kpCorrection)[j]$ denote the $i$-th and $j$-th points in the input \inputPC and posed CAD model $\predCAD(\kpCorrection) = \MT(\kpCorrection) \cdot \cadPC$, respectively.  
Here, \cadPC denotes a dense sampling of points on the CAD model \cad, and $\rho(\cdot)$ is a truncated  
least squares (TLS) robust loss function defined as $\rho(z) = \min\{z^2, \bar{c}^2\}$, where $\bar{c}$ is a user-specified threshold. 
In words, problem~\eqref{eq:corrector} computes the optimal keypoint correction such that the corrected keypoints produce a pose estimate $\MT(\kpCorrection)$  (via the last constraint in~\eqref{eq:corrector}) such that the CAD model posed at $\MT(\kpCorrection)$ (second constraint in~\eqref{eq:corrector}) matches the input (objective in~\eqref{eq:corrector}).
The objective minimizes the distance between point pairs with distance below $\bar{c}$ and disregards points in the input $\inputPC$ for which the closest point  in $\predCAD(\kpCorrection)$ is at a distance farther than $\bar{c}$.
We use the TLS loss to automatically reject outliers in the input \inputPC, when solving the corrector optimization problem. The TLS loss is commonly used in robust estimation for robotics~\cite{Yang20ral-GNC,MacTavish15crv-robustEstimation,Yang20tro-teaser} due to its insensitivity to  outliers.

\myParagraph{Forward Pass} The bi-level optimization problem~\eqref{eq:corrector} is non-linear and non-convex. However, we note that the lower-level optimization problem (last line in~\eqref{eq:corrector}) is nothing but an outlier-free registration problem, which can be solved in closed form via singular value decomposition (SVD)~\cite{Horn88, Arun87pami}. Since the SVD computation is differentiable, we can directly compute the gradient of the cost function with respect to the correction $\kpCorrection$. 
Therefore, we can solve the robust corrector problem~\eqref{eq:corrector} via a constant-step-size gradient descent.
In our experiments, we implement the solver in PyTorch~\cite{Paszke19-pytorch} ---so as to enable batch processing--- 
and compute the gradients using the autograd functionality in PyTorch.

\myParagraph{Backward Pass} In order to include the robust corrector as a module of a trainable 
architecture, it is necessary to differentiate through the corrector optimization problem. We now show that the robust corrector optimization problem, although non-linear and non-convex, admits a very simple derivative of the output \kpOptCorrection, with respect to the input \detKP.

\begin{proposition}
	The gradient of the correction \kpOptCorrection with respect to the detected keypoints \detKP is the negative identity:
	\begin{equation}
	\partial \kpOptCorrection / \partial \detKP = -\mathbf{I}.
	\end{equation}
\end{proposition}

We omit the proof of the proposition, which remains identical to its non-robust counterpart in~\cite{Talak23arxiv-c3po}. 

\begin{remark}[Robust Corrector vs. Non-Robust Corrector vs. ICP]
The work~\cite{Talak23arxiv-c3po} proposes a non-robust version of the corrector, that is recovered by using $\rho(z) = z^2$ in~\eqref{eq:corrector}. The work also showed the advantage of optimizing over keypoints, rather than over poses (\eg as in ICP).
As we show in our experiments, the non-robust loss in~\cite{Talak23arxiv-c3po} leads to large errors when the input point cloud contains outliers resulting from a noisy 2D segmentation. On the other hand, the proposed approach is able to correct large errors even in the presence of a large fraction of outliers. %
\end{remark} %

\begin{figure*}[t!]
	\centering 
	\includegraphics[trim=60 70 120 10,clip,width=\linewidth]{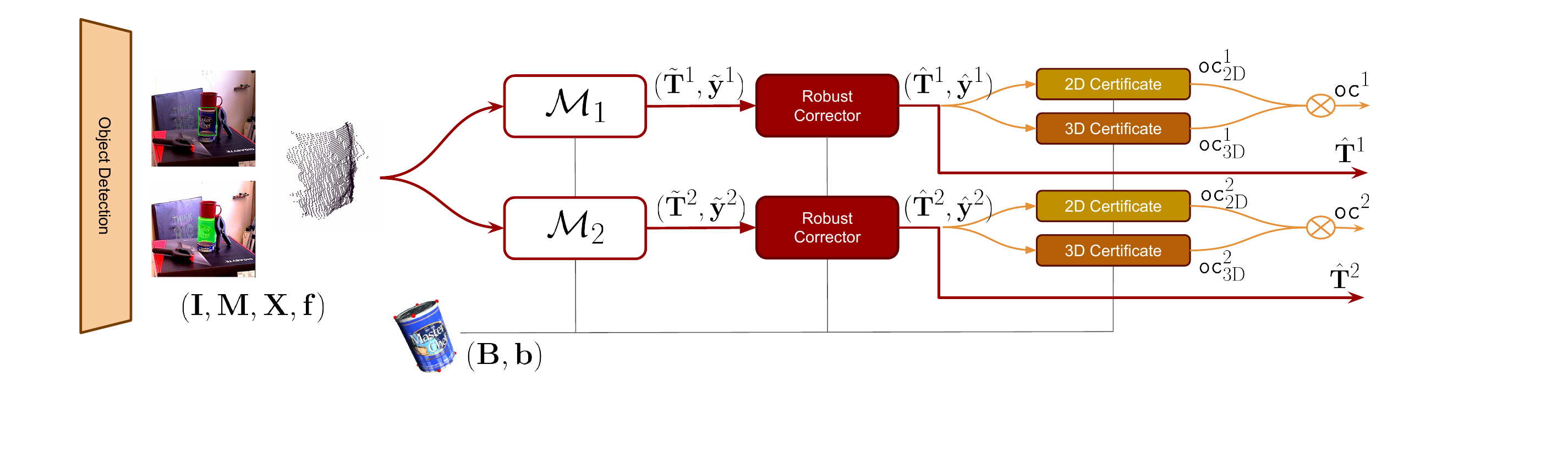}
	\caption{\textbf{\Es architecture.} The proposed architecture stacks several pose estimation models in parallel ---we show only two for the sake of simplicity: model \Mone and \Mtwo. The models take in the outputs produced by a 2D object detection and segmentation: \inputRGBD (scaled RoI of the detected object), \inputMask (2D segmentation mask of the detected object), \inputPC and \inputFeat (point cloud of the detected object and color as point feature). The two models output object pose estimates and keypoints: \outOne, \outTwo. The \emph{robust corrector}, corrects the estimated poses, using dense information from the input \inputPC and the object CAD model \cad. It produces corrected pose and keypoints: \correctedOne, \correctedTwo. Two binary certificates, namely, the 2D and 3D certificates check geometric consistency of the corrected pose \estPoseOne and \estPoseTwo. If the 2D and 3D checks succeed, the corrected pose is declared to be \emph{observably correct}. 
	The architecture outputs an observably correct pose, and if none is found, it outputs the corrected pose of a preferred model, say pose \estPoseOne from \Mone. \vspace{-4mm}}
	\label{fig:parallel-arch}
\end{figure*}

\section{\ES Architecture}
\label{sec:parallel}

\myParagraph{Overview}
\Cref{fig:parallel-arch} illustrates the proposed \es architecture. In the figure and description below, we limit the scope to stacking only two pose estimation models in parallel, but the methodology presented here can be trivially extended to a larger number of models (see Remark~\ref{rmk:22Many} below). 

Let \Mone and \Mtwo be two pose estimators that take in \inpuT and output estimated pose and detected keypoints; below we will observe that keypoints detections can be artificially added to any pose estimator, as long as we have a CAD model of the object.
We use \outOne and \outTwo to denote the outputs produced by models \Mone and \Mtwo, respectively. 
We augment each pose estimators with two modules: (i) robust corrector and (ii) observable correctness certificate check. The robust corrector takes in the estimated pose and detected keypoints $(\detPose, \detKP)$, and outputs corrected pose and keypoints $(\estPose^i, \predKP^i)$, $i=1,2$, see~\cref{sec:robustCorrector}. 

The observable correctness certificate implements two checks to assess the geometric consistency of the corrected pose. The 3D certificate checks if the input point cloud \inputPC matches the posed CAD model $\hat{\MT}^i\cdot \cad$, for $i=1,2$. The 2D certificate, on the other hand, checks if the rendering of the posed CAD model, namely $\hat{\MT}^i\cdot \cad$, on the image plane is compatible with the input segmentation mask \inputMask. Both tests account for occlusions and outliers in the detected object.
These two certificates in turn determine if the corrected poses $\hat{\MT}^1$ and $\hat{\MT}^2$, produced by the robust correctors, are observably correct or not. The self-training uses the observably correct instances to supervise the training process.
We now describe each of these modules, and the self-training method, in detail.

\subsection{Robust Corrector Beyond Keypoint-based Models}
\label{sec:corrector}
In~\cref{sec:robustCorrector}, we observed that a semantic-keypoint-based pose estimator 
processes the sensor data to obtain keypoint detections $\detKP$ and pose estimate $\detPose$.
Therefore, we can think about such an estimator as a map from the input $(\inputRGBD, \inputMask, \inputPC, \inputFeat)$ to the tuple $(\detKP, \detPose)$ of detected keypoints and pose estimate:
\begin{equation}
\label{eq:pe-mapping}
\calM: (\inputRGBD, \inputMask, \inputPC, \inputFeat) \rightarrow (\detKP, \detPose),
\end{equation}
where for the sake of generality we let $\calM$ take \inpuT as input, whereas an estimator may use 
only a subset of the four inputs (\eg a point-cloud-only pose estimator may choose to work with only \inputPC, in the four-tuple).

It is now easy to realize that if a pose estimator does not rely on keypoints and only computes a pose estimate $\detPose$, we can still hallucinate semantic-keypoint detections as 
\begin{equation}
\label{eq:kp-from-pose}
\detKP = \detPose \cdot \kp,
\end{equation}
where $\kp$ are the keypoints on the CAD model. 
Using $\detKP$ as keypoint detections, we can then apply the robust corrector optimization problem~\eqref{eq:corrector} to any pose estimator, as long as we have a CAD model $\cad$ of the object we aim to detect.

\subsection{Certificate of Observable Correctness}
\label{sec:certificates}
This section describes binary certificates that can distinguish correct estimates from incorrect ones, \ie a certificate that is equal to one when the estimate from a pose estimator is correct and zero otherwise.
More precisely, we design certificates of \emph{observable correctness}~\cite{Talak23arxiv-c3po} that determines if the output produced by a learning-based model is consistent with the input data and the object CAD model.
We propose two certificates of observable correctness, that assess consistency with both the 2D segmentation mask and the 3D point cloud.

\myParagraph{3D Certificate}
Let \estPose be the corrected pose produced by the robust corrector.  
We compute the posed CAD model $\predCAD = \estPose \cdot \cadPC$,
posed at \estPose. Consider the set of distance-scores: 
\begin{equation}
\scoreSet{\inputPC}{\predCAD} = \{ s_i = \min_{j} \lVert \inputPC[i] -\predCAD[j] \rVert_{2}~|~i=1,\ldots,n\}.
\end{equation}
The 3D certificate is then given by
\begin{equation}
\label{eq:cert3d}
\ocThreeD{\inputPC}{\estPose} = \indicator{
{\tt percentile}( \scoreSet{\inputPC}{\predCAD}, p)
< \epsilon_{\texttt{\cThreeD}}},
\end{equation}
where ${\tt percentile}(\scoreSet{\inputPC}{\predCAD}, p)$ is the $p$-th percentile of the vector $\scoreSet{\inputPC}{\predCAD}$ ($p=0.9$ in our tests), and
 $\epsilon_{\texttt{\cThreeD}}$ is a user-specified threshold.
Intuitively, for $p=0.9$,  $\ocThreeD{\inputPC}{\estPose} = 1$ if at least 90\% of the points in $\inputPC$ is within a distance $\epsilon_{\texttt{\cThreeD}}$ from a point in \predCAD, or $\ocThreeD{\inputPC}{\estPose} = 0$ otherwise. The percentile is used to account for outliers: taking a 90\% percentile corresponds to assuming that 
the outliers $\outliersPC$ in~\eqref{eq:inputPCgen} do not corrupt more than 10\% of the points in the input $\inputPC$. 

\myParagraph{2D Certificate}
Let \estPose be the corrected pose produced by the robust corrector.  
We render the posed CAD model $\estPose \cdot \cad$ to obtain the mask $\estMask$. 
Note that the \estMask will render the entire posed CAD model, hence will not account for occlusions, caused by other objects.
In order to ensure that our 2D certificate is occlusion-aware, we compare the area of the detected mask \inputMask with the area of the intersection $\inputMask \cap \estMask$. 
The 2D certificate is then given by 
\begin{equation}
\label{eq:cert2d}
\ocTwoD{\inputPC}{\estPose} = \indicator{ \frac{\area{\inputMask \cap \estMask}}{\area{\inputMask}} > 1 - \epsilon_{\texttt{\cTwoD}}},
\end{equation}
where $\area{\inputMask}$ denotes the pixel area of all pixels $(i, j)$ in the mask \inputMask with $\inputMask(i, j) = 1$, and $\epsilon_{\texttt{\cTwoD}}$ is a given threshold.

\myParagraph{Observable Correctness Certificate}
We say that an output \correctedOutput is \emph{observably correct} if the two checks ---the 3D and 2D certificates--- are met. More formally, we deem an output $\correctedOutput$ to be observably correct when $\oc{\inputPC}{\estPose} = 1$ with:
\begin{equation}
\oc{\inputPC}{\estPose} = \ocTwoD{\inputPC}{\estPose} \cdot \ocThreeD{\inputPC}{\estPose}.
\end{equation}
In the next subsection, we show how to use the binary certificates to 
self-supervise an \ensemble of pose estimators.

\begin{remark}[Certificates vs Certifiable Correctness]
The work~\cite{Talak23arxiv-c3po} shows that one can formally guarantee the correctness of a pose estimate (\ie its proximity to the ground truth) by computing a certificate of observable correctness 
and a certificate of non-degeneracy. The former evaluates how well the estimate fits the data; the latter ensures that the data contains enough information to compute a unique estimate. 
In this work, we only define certificates of observable correctness, hence we cannot derive formal performance guarantees. This is mostly due to the fact that we consider a realistic case with noisy segmentation, while the analysis~\cite{Talak23arxiv-c3po} assumes perfect segmentation. 
Despite this theoretical gap, we empirically observe that the certificates of observable correctness 
provide an excellent tool for identifying good estimates.
\end{remark}

\subsection{\ES}
\label{sec:training}
We now describe our \es procedure that trains multiple models in parallel, using just the binary-valued, observable correctness certificate to supervise. 

\myParagraph{Loss Functions} We first define two loss functions involved in our \es. For a corrected \estPose and input $\inputPC$, we define the self-supervised loss to be:
\newcommand{\lossSelf}[2]{\ensuremath{\calL_{\text{self}}\left( #1, #2\right)}\xspace}
\newcommand{\lossSup}[2]{\ensuremath{\calL_{\text{sup}}\left( #1, #2\right)}\xspace}
\newcommand{\nrPointsCAD}{\ensuremath{m}\xspace}
\begin{equation}
\label{eq:loss-self}
\lossSelf{\inputPC}{\estPose} = \frac{1}{n}\sum_{i}\rho\left( \min_{j} \lVert \inputPC[i] - \estPose\cdot \cadPC[j] \rVert_{2}\right),
\end{equation}
where $\rho(z) = \min\{z^2, \bar{c}^2\}$ and $\bar{c}$ denotes the maximum admissible distance between pairs of points to be considered inliers.
For a pose $\MT'$, we also define a supervised loss to be
\begin{multline}
\label{eq:loss-sup}
\lossSup{\estPose}{\MT'} = \frac{1}{\nrPointsCAD}\sum_{i} \min_{j} \lVert \estPose\cdot\cadPC [i] - \MT' \cdot \cadPC[j] \rVert^{2}_{2} \\
+ \frac{1}{m}\sum_{j} \min_{i} \lVert \estPose\cdot\cadPC [i] - \MT' \cdot \cadPC[j] \rVert^{2}_{2},
\end{multline}
where $\nrPointsCAD$ denotes the number of points on the sampled point cloud \cadPC. 
In our definition in~\eqref{eq:loss-sup}, we assume the loss to only back-propagate through \estPose (to train model weights), and not through $\MT'$.
We remark that the supervised loss is nothing but the ADD-S~\cite{Wang19cvpr-DenseFusion6D} loss between the two posed models, namely $\estPose\cdot\cadPC$ and $\MT' \cdot \cadPC$; below we are going to use it in a way that each model in the \ensemble supervises the others.

 \myParagraph{Self-Training}
Recall that \estPoseOne and \estPoseTwo denote the two poses outputted by model \Mone and \Mtwo and refined by the robust corrector (\cref{fig:parallel-arch}). 
Our self-training relies on stochastic gradient descent (SGD). 
At each SDG iteration and for each input \inputPC in a batch, we induce the following training loss:
\begin{multline}
\label{eq:loss-train}
\calL = \oc{\inputPC}{\estPoseOne} \cdot \left[ \lossSelf{\inputPC}{\estPoseOne} + \lossSup{\estPoseTwo}{\estPoseOne} \right] \\
+ \oc{\inputPC}{\estPoseTwo} \cdot \left[ \lossSelf{\inputPC}{\estPoseTwo}  + \lossSup{\estPoseOne}{\estPoseTwo} \right].
\end{multline}
The supervised loss ${\lossSup{\estPoseTwo}{\estPoseOne}}$ only back-propagates through $\estPoseTwo$, and trains model \Mtwo, and not \Mone. Similarly, $\lossSup{\estPoseOne}{\estPoseTwo}$ back-propagates through $\estPoseOne$, and trains model \Mone, and not \Mtwo.
Note that the loss only uses the input and the output of the different models but does not rely on external supervision.
 $\oc{\inputPC}{\estPoseOne}$ and $\oc{\inputPC}{\estPoseTwo}$ are the binary-valued, observable correctness certificates. For instance, $\oc{\inputPC}{\estPoseOne} = 1$ indicates that the pose $\estPoseOne$ is observably correct.
Therefore, whenever \estPoseOne is observably correct,
the loss~\eqref{eq:loss-train} induces a self-supervised loss $\lossSelf{\inputPC}{\estPoseOne}$ on model \Mone and a supervised loss $\lossSup{\estPoseTwo}{\estPoseOne}$ on model \Mtwo (with supervision using \estPoseOne). Similarly, 
 it induces a loss on \Mone and \Mtwo, when \estPoseTwo is observably correct. 
This ensures that the observably correct outputs produced by one model are used to train the second model, and vice versa, taking advantage of the potential complementarity of the models in the \ensemble. 

\begin{remark}[From Two to Many]\label{rmk:22Many}
The training loss~\eqref{eq:loss-train} is specified for training two models in parallel. This loss function can easily be extended to simultaneously train $K$ pose estimation models. Let $\estPoseOne, \estPoseTwo, \ldots \estPose^{K}$ denote the corrected poses output by the $K$ pose estimators, after applying  the robust corrector to each model. The training loss of our \es ---for each input \inputPC--- would then be:
\begin{multline}
\sum_{k=1}^{K} \oc{\inputPC}{\estPose^{k}}
\cdot \left[ \lossSelf{\inputPC}{\estPose^{k}} + \sum_{k'\neq k} \lossSup{\estPose^{k'}}{\estPose^{k}}\right].\nonumber 
\end{multline}
Recall that the loss $\lossSup{\estPose^{k'}}{\estPose^{k}}$ only back-propagates  through $\estPose^{k'}$ to train model $\calM^{k'}$, and not $\calM^{k}$. %
\end{remark}

\begin{remark}[Role of the Robust Corrector]
The robust corrector is instrumental in
extending the reach of each trained model. That is, it retains high accuracy for a much larger set of inputs, than what the pose estimation model ---without the corrector--- initially does.
This allows bootstrapping the self-training process, \ie ensures that during the initial self-training iterations there are enough observably correct inputs (which are then used for self-training) despite a potentially large domain gap; see results in~\cref{sec:ycbv-tless-expt}.
\end{remark}

\section{Outlier-Robust Point Cloud Processing}
\label{sec:rkeypo}

This section describes two improvements to existing regression-based architectures for keypoint detection; 
these improvements are mostly minor, but impact performance in practice, and are broadly applicable to a number of architectures, including point transformers~\cite{Zhao21iccv-PointTransformer} and point-net~\cite{Qi17cvpr-pointnet}.

\myParagraph{Challenges of Regression-based Architectures}
The unstructured nature of point clouds makes it harder to device a simple pooling layer, that simultaneously samples representative points in the point clouds, and aggregates nearby features. Most point cloud-based architectures have resorted to using farthest point sampling strategy (FPS)~\cite{Qi17cvpr-pointnet, Qi17nips-pointnet++, Zhao21iccv-PointTransformer}, followed by aggregating features from the k-nearest neighbors. Such a pooling layer assumes that maximizing the distance between the sampled points is a good heuristic to select representative points in the point cloud.
However, in the presence of outliers, FPS ---by construction--- tends to sample many outliers, instead of rejecting them. 
Furthermore, the lack of intrinsic translation invariance/equivariance also renders point-cloud-based models sensitive to the choice of a centroid, which is used in practice to center the point cloud and  regain partial translation invariance/equivariance.
To address these issues, we propose a \emph{robust centroid} computation and \emph{robust pooling}, 
which enhance the robustness of point-cloud-based models in the presence of outliers in the input points.  

\myParagraph{Robust Centroid}
In the presence of outliers, the algebraic mean of the points in the point cloud $\inputPC$ might be a poor proxy for the object center. We proposed to compute a \emph{robust centroid} by solving a robust estimation problem:
\begin{equation}
\label{eq:prob:robust-centroid}
\bar{\vxx} = \argmin_{\vu \in \Real{3}} \frac{1}{\nrPoints} \sum_{i} \rho\left( \lVert \inputPC[i] - \vu\rVert_{2} \right),
\end{equation}
where $\rho(\cdot)$ is a robust loss function (\eg TLS). 
We solve~\eqref{eq:prob:robust-centroid} using the graduated non-convexity algorithm~\cite{Yang20ral-GNC}. 

The input point cloud is then centered at $\bar{\vxx}$, namely,
\begin{equation}
\inputPC \leftarrow \inputPC - \bar{\vxx}\kron \ones_{\nrPoints}\tran 
\end{equation}
where $\ones_{\nrPoints}$ is the vector of ones, before passing it to the point-cloud architecture. If the output is desired to be translation equivariant, the robust centroid $\bar{\vxx}$ is also added to the model output.
We remark that the robust centroid computation does not require to be differentiable, 
since it is applied directly to the input and there are no trainable weights in~\eqref{eq:prob:robust-centroid}. %

\myParagraph{Robust Pooling}
Pooling is a building block of existing keypoint detection architectures, see~\cite{Qi17cvpr-pointnet, Qi17nips-pointnet++, Zhao21iccv-PointTransformer}.
We propose a simple, trainable pooling layer, which samples points based on regressed scores for each point.
The pooling layer takes an input point cloud with features $(\inputPC, \inputFeat) \!\in\! \Real{3\times \nrPoints} \!\times\! \Real{d\times \nrPoints}$ and outputs another point cloud, with features $(\inputPC', \inputFeat')  \!\!\in\! \Real{3\times n'} \!\times\!  \Real{d\times n'}$; $d$ denotes the feature dimension ($d\!=\!3$ in our pose estimation setup), and $n' \!<\! \nrPoints$ since the number of pooled points is always smaller than the number of input points.

\JS{Given the input features $\inputFeat = [\vf_1, \ldots \vf_\nrPoints]$, we first regress a score $s_i =\mlp{\vf_i}$ for each point $i$ using a multilayer perceptron \mlp{\cdot}. We select $\nrPoints'$ ($< \nrPoints$) points in $(\inputPC, \inputFeat)$ that have the highest scores to obtain $(\inputPC', \inputFeat')$.
In our experiments, we implement \mlp{\cdot} to have one hidden layer and ReLU activations.
The trainable \mlp{\cdot} weights make the layer malleable during training to automatically learn to reject outliers in the input point cloud. 
}

\begin{remark}[Robust Pooling as Trainable Sampling]
The robust pooling layer only samples points and does not aggregate features from nearby points to create new features --- as a traditional pooling layer does. We assume here that other blocks in the architecture will learn to extract the necessary features, before pooling. Treating pooling as trainable sampling not only simplifies the architecture, but also saves compute time that is otherwise spent in performing k-nearest-neighbor search and aggregation operation.
\end{remark}

\begin{figure*}
	\centering 
	\begin{subfigure}{0.32\textwidth}
		\includegraphics[trim=0 0 0 0,clip,width=\linewidth]{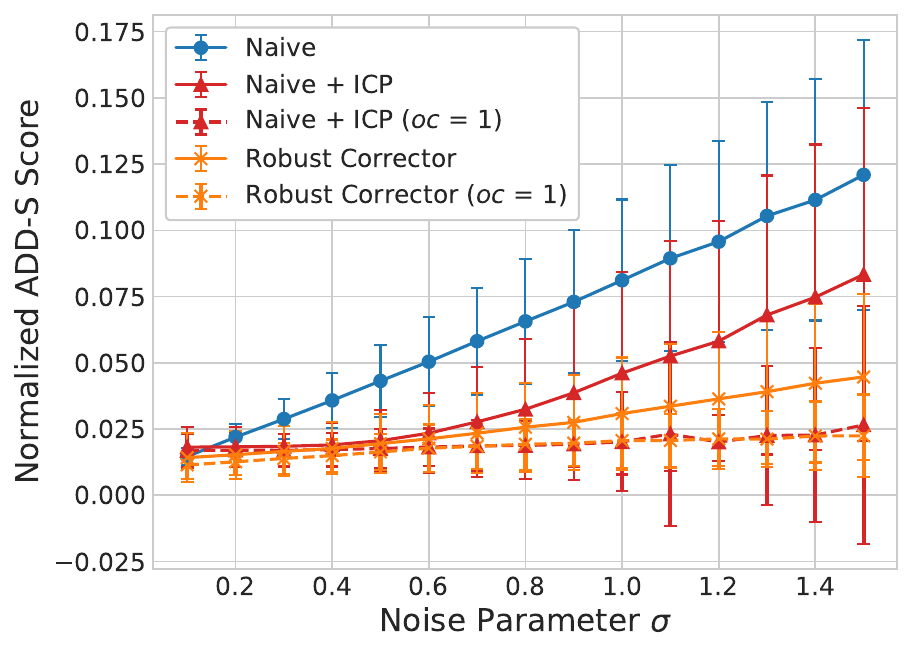}
				\caption{ \label{fig:corrector-analysis-ycbv-adds} }
	\end{subfigure}
	\hspace{3mm}
	\begin{subfigure}{0.3\textwidth}
		\includegraphics[trim=0 0 0 0,clip,width=\linewidth]{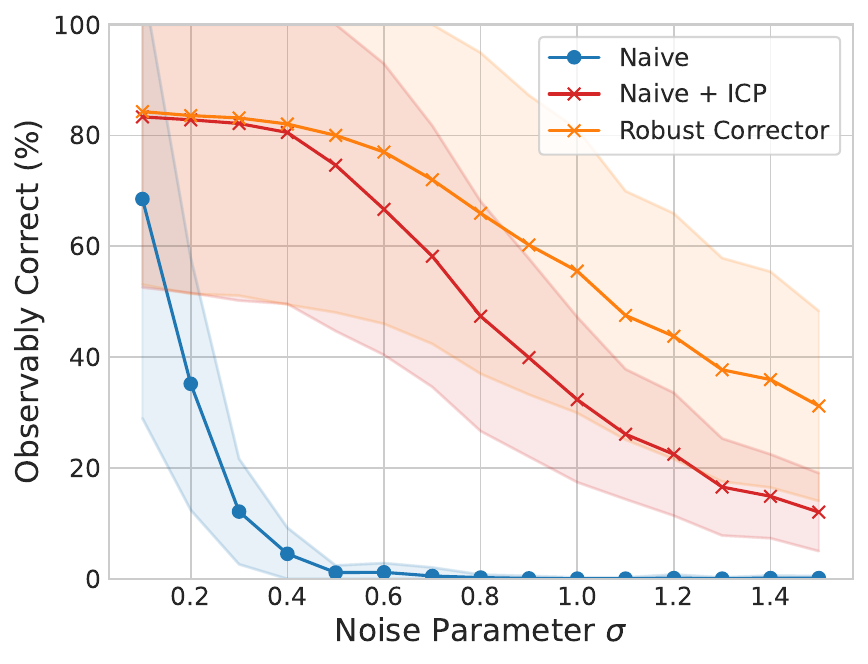}
				\caption{  \label{fig:corrector-analysis-ycbv-cert} }
	\end{subfigure}
	\hspace{3mm}
	\begin{subfigure}{0.27\textwidth}
		\includegraphics[trim=0 0 0 0,clip,width=\linewidth]{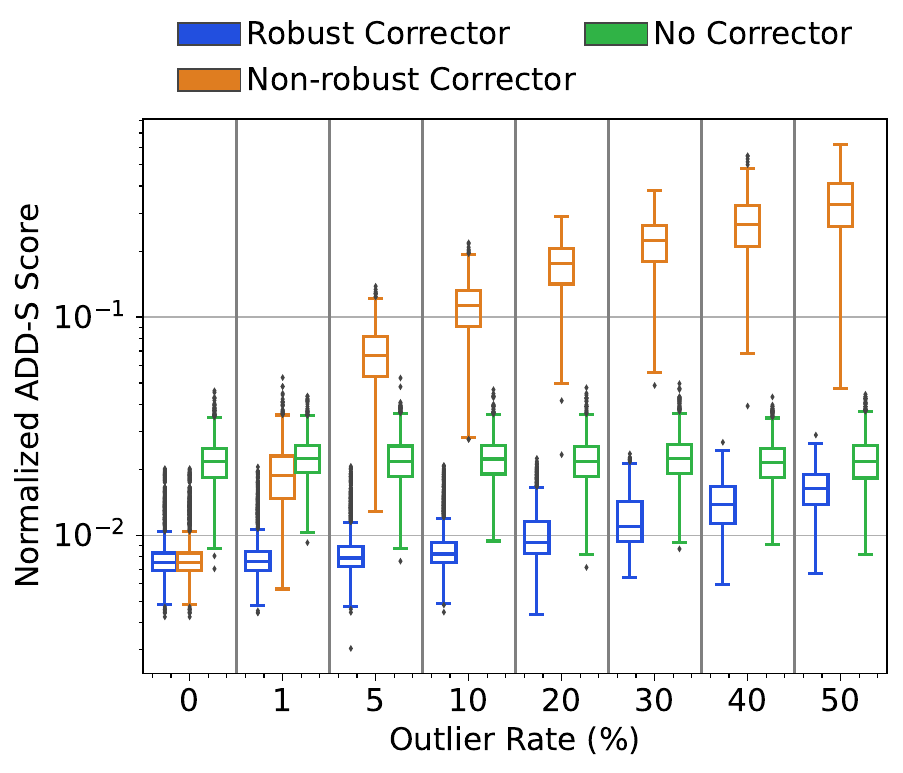}
				\caption{ \label{fig:robust-vs-non-robust}  }
		
	\end{subfigure}
	\caption{\JS{(a) Normalized ADD-S (averaged across all objects in the \YCBV dataset) and corresponding standard deviation (shown as error bars) as a function of the noise parameter $\sigma$. (b) Fraction of observably correct instances (\ocx = 1) (averaged across all objects in the \YCBV dataset) as a function the noise parameter $\sigma$. (c) ADD-S as a function of the outlier rate.}}%
	\label{fig:corrector-analysis}
\end{figure*}

\section{Experiments}
\label{sec:expt}

We present four sets of experiments. We first demonstrate the effectiveness of the robust corrector in correcting large keypoint errors (\cref{sec:corrector-analysis}), and show its utility  viz-a-viz the non-robust corrector proposed in~\cite{Talak23arxiv-c3po} (\cref{sec:robust-vs-non-robust}). We then show the ability of the robust centroid and pooling strategies, developed in \cref{sec:rkeypo}, in mitigating the effects of outliers during keypoint detection (\cref{sec:rkeypo}). Finally, we show the effectiveness of our \es architecture by self-training a point-cloud-based model and a CNN-based model (CosyPose~\cite{Labbe20eccv-CosyPose}) in parallel (\cref{sec:ycbv-tless-expt}).

\subsection{Robust Corrector Analysis}
\label{sec:corrector-analysis}

\myParagraph{Setup} We use the \YCBV dataset and objects~\cite{Xiang17rss-posecnn, Hodan20eccvw-BOPChallenge}. For each \YCBV object, we extract the object depth point cloud from the \YCBV test set. In this subsection, we use the ground-truth pose annotations to get the ground-truth keypoint annotations using~\eqref{eq:kp-from-pose}, and we set the detected keypoints \detKP to be ground-truth semantic keypoints plus an additive perturbation.
 If \gtKP denotes the ground-truth keypoints, our noise model adds uniform noise in range $[-\sigma D/2, \sigma D/2]$, to keypoint $\gtKP[i]$, with probability $f$, where $D$ is the object diameter; this allows us to simulate increasing levels of keypoint detection errors by increasing $\sigma$. We set $f=0.8$ and analyze the robust corrector as a function of the noise variance parameter $\sigma$. 

\myParagraph{Results} 
The robust corrector is designed to correct keypoint detection errors.
\cref{fig:corrector-analysis}(a) plots the normalized ADD-S (\ie the ADD-S normalized by the object diameter $D$) as a function of the keypoint detection noise parameter $\sigma$. We plot these scores for the robust corrector, as well as the observably correct instances produced by the robust corrector.
\JS{We evaluate against two baselines: (i) \naive: which sets $\predKP = \detKP$ in solving the outlier-free registration problem~\eqref{eq:corrected-pose}, and (ii) \naiveICP: which uses outlier-free registration (same as \naive) and then refines the result with point-to-point ICP.
The threshold $\bar{c}$ for the robust corrector and the maximum points-pair correspondence distance in ICP are both set to be $30\%$ of the object diameter.
\cref{fig:corrector-analysis}(b) plots the fraction of observably correct instances produced by each method, as a function of~$\sigma$.\hspace{-5mm}
}

\myParagraph{Insights}
We observe that the robust corrector is able to correct large keypoint errors, leading to improved pose estimates (\cref{fig:corrector-analysis}(a)). Even when $\sigma = 0.6$, \ie when $80\%$ of the keypoints are perturbed by uniform noise proportional to $60\%$ of the object diameter, we see that \JS{around $80\%$} of the outputs produced by the robust corrector are observably correct (\cref{fig:corrector-analysis}(b)), and the observably correct instances are highly accurate (dashed red line in~\cref{fig:corrector-analysis}(a)). This number drops to $0\%$ for \naive (\cref{fig:corrector-analysis}(b)), which indeed exhibits much larger errors (\cref{fig:corrector-analysis}(a)).
\JS{\naiveICP outperforms \naive, in terms of accuracy and fraction of observably correct instances. The robust corrector outperforms both --- especially for large noise ($\sigma$) values.}

\subsection{Robust versus Non-Robust Corrector}
\label{sec:robust-vs-non-robust}

\myParagraph{Setup} We implement the non-robust corrector proposed in~\cite{Talak23arxiv-c3po}, which solves~\eqref{eq:corrector}, but with $\rho(z) = z^2$. We compare the latter against the proposed robust corrector in the presence of increasing number of outlier points in the input \inputPC. We use a setup similar to \cref{sec:corrector-analysis}, but now we fix $\sigma=0.4$ (along with $f=0.8$), and add outliers to \inputPC. {We evaluate the ADD-S score as a function of the outlier rate: for instance, when the outlier rate is 0.5, we test on a point cloud where 50\% of the points have been replaced with random points.}

\myParagraph{Results and Insights} \Cref{fig:corrector-analysis}(c) plots the normalized ADD-S score
as a function of the outlier rate. We observe that the proposed robust corrector significantly outperforms the non-robust corrector from~\cite{Talak23arxiv-c3po}. We also observe that, in the presence of outliers, the non-robust corrector yields worse performance than having no corrector at all, which is expected, since the non-robust corrector will be increasingly biased by the outliers in the point cloud \inputPC.

\subsection{Impact of Robust Centroid and Robust Pooling}
\label{sec:robust-cnp}
\begin{figure}
	\centering
	\includegraphics[width=0.85\linewidth]{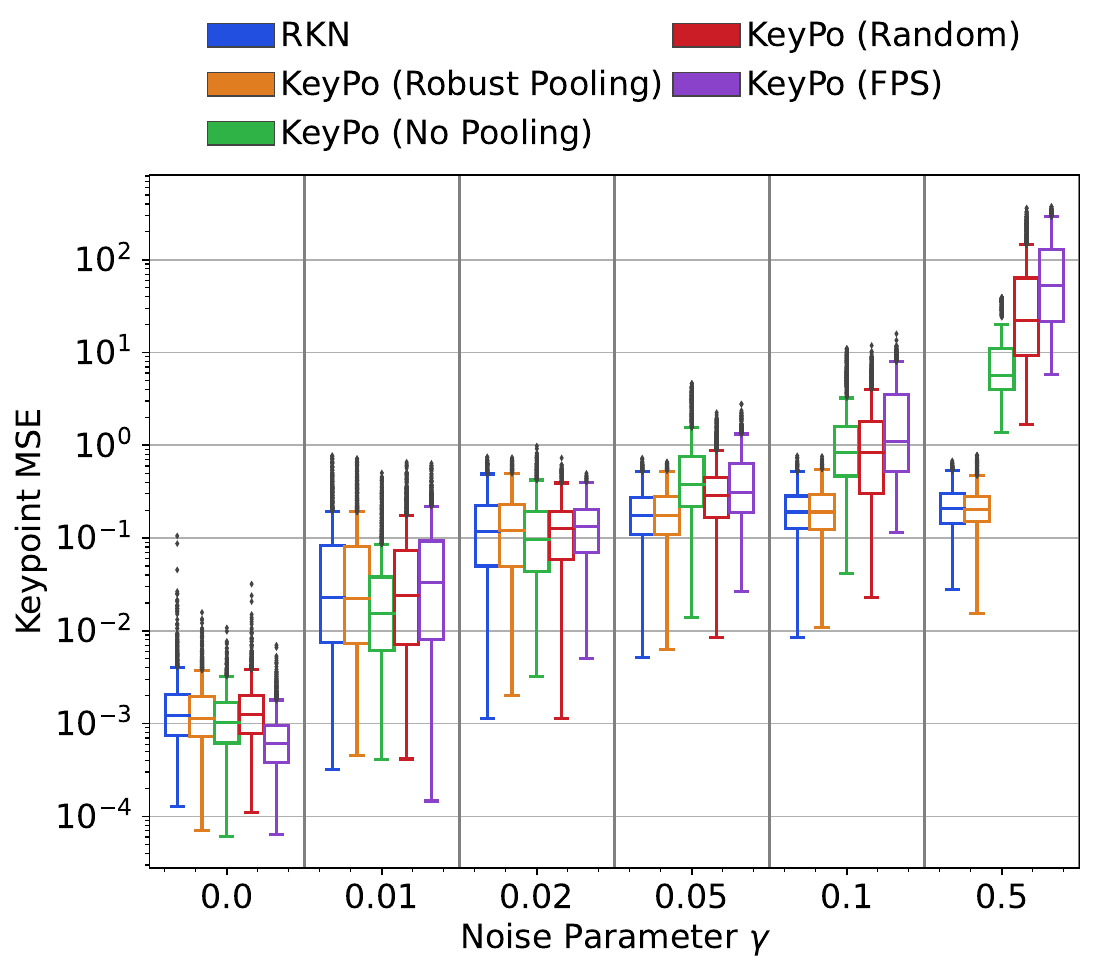}
	\caption{Mean squared error (MSE) in the keypoint detection as a function of the added noise variance $\gamma$.}
	\label{fig:robust-pc-noise}
\end{figure}

\begin{figure}
	\centering
	\includegraphics[width=0.85\linewidth]{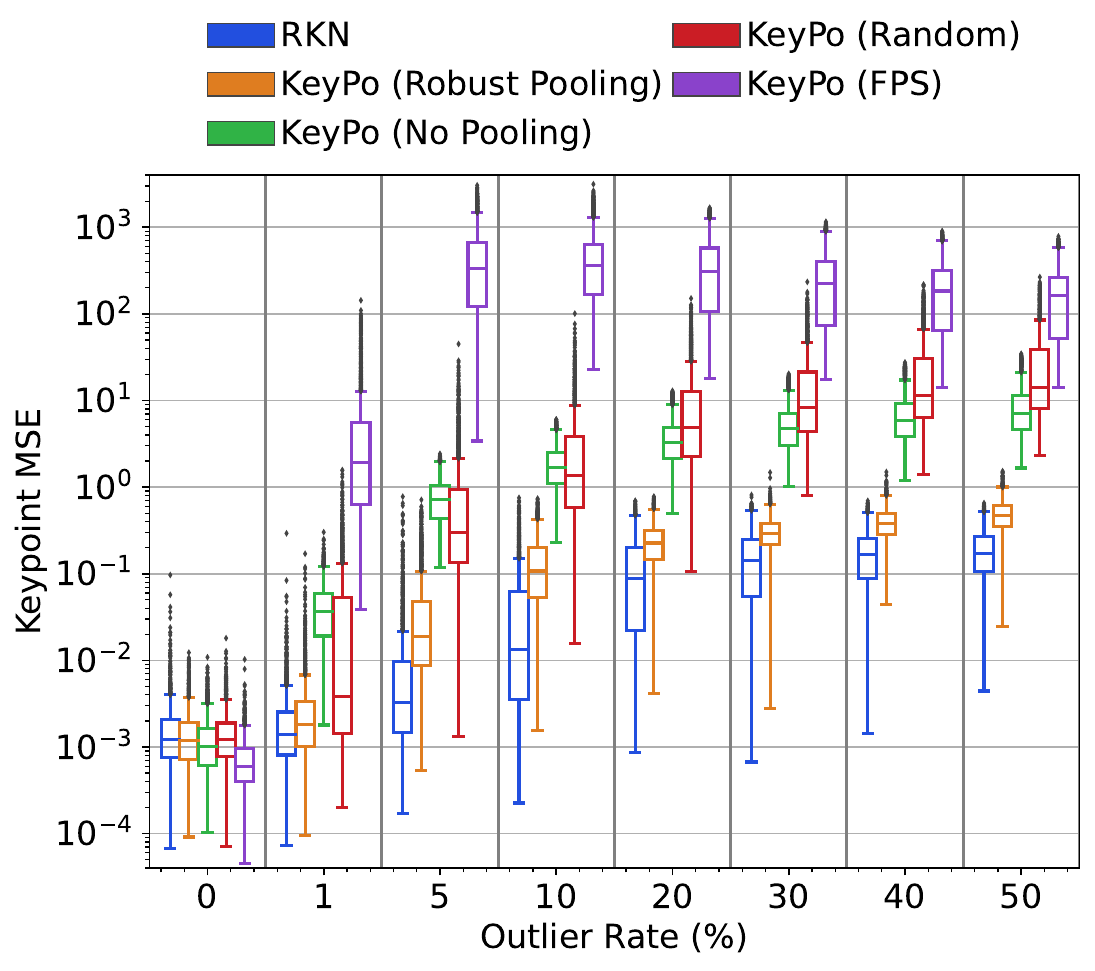}
	\caption{Mean squared error (MSE) in the keypoint detection as a function of the outlier rate in the input \inputPC.}
	\label{fig:robust-pc-outliers}
\end{figure}

\myParagraph{\rKeyPoLong (\rKeyPo)} To evaluate the effect of the robust centroid and robust pooling, we integrate these two modules into a point transformer architecture~\cite{Zhao21iccv-PointTransformer} for semantic keypoint detection. In particular, we subtract the robust centroid to the point cloud before passing it to the detector. 
Then, in the architecture, we alternate each point transformer block~\cite{Zhao21iccv-PointTransformer} 
(which extracts and transforms features), with a robust pooling layer (that sparsifies the point cloud). 
This is repeated several times, and the final result is passed to a multi-layered perceptron, which regresses the keypoints \detKP. Finally, the centroid is added back to the keypoints to regain translation 
equivariance. We call the resulting architecture \emph{\rKeyPoLong} (\rKeyPo): \rKeyPo regresses semantic keypoints \detKP, given a point cloud with color as point features.

\myParagraph{Setup}
We evaluate \rKeyPo against baselines that use the same architecture as \rKeyPo, but without the robust pooling and centroid; we refer to those with the label ``\KeyPo''. %
In particular, we consider the following variants: (i) \KeyPo (FPS): which is \KeyPo with farthest point sampling for pooling, (ii) \KeyPo (Random): which is \KeyPo with random point sampling for pooling, (iii) \KeyPo (No Pooling): which is \KeyPo without any pooling layers, (iv) \KeyPo (Robust Pooling): which is \KeyPo with the proposed robust pooling, and finally, (v) \rKeyPo, which is nothing but \KeyPo with the proposed robust centroid and robust pooling. We remark that \KeyPo (FPS), \KeyPo (Random), \KeyPo (No Pooling), and \KeyPo (Robust Pooling) use a non-robust centroid computation. %
We use the \YCBV dataset~\cite{Xiang17rss-posecnn, Hodan20eccvw-BOPChallenge} and train the keypoint detectors in a fully supervised manner. 
We evaluate the trained models in two settings. The first adds zero-mean Gaussian noise, with standard deviation $\gamma$, to each point in the input \inputPC. The second adds outlier points to the input \inputPC. 

\myParagraph{Results and Insights} 
\cref{fig:robust-pc-noise} plots the mean squared error (MSE) in the detected keypoints as a function of the added noise standard deviation $\gamma$;  
\cref{fig:robust-pc-outliers} 
plots the MSE of the detected keypoints as a function of the outlier rate in \inputPC.
 We observe that \rKeyPo, with robust centroid and robust pooling, outperforms all the baselines. We see that while \KeyPo (Robust Pooling) shows performance competitive to \rKeyPo in the case of added noise but no outliers (\cref{fig:robust-pc-noise}), it does not fare well in the presence of outliers. We also observe that popular pooling methods, such as farthest point sampling and random sampling, tend to do worse than using no sampling at all.

\subsection{The \YCBV and \TLESS Experiment}
\label{sec:ycbv-tless-expt}

\begin{figure*}
	\centering 
	\begin{subfigure}{0.33\textwidth}
		\includegraphics[trim=0 0 0 0,clip,width=\linewidth]{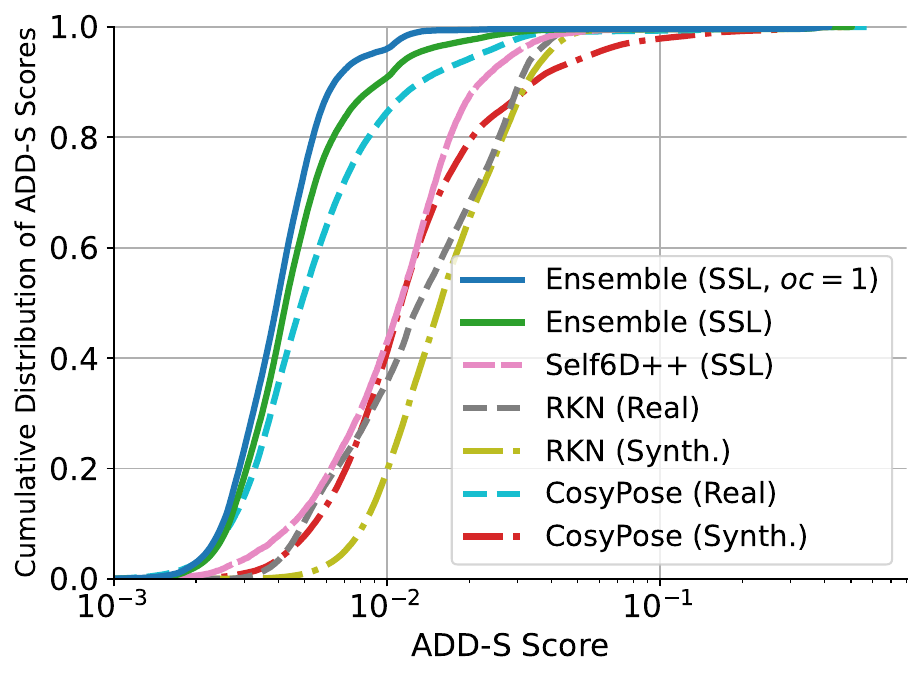}
				\caption{ }
		\label{fig:baseline}
	\end{subfigure}
	\begin{subfigure}{0.33\textwidth}
		\includegraphics[trim=0 0 0 0,clip,width=\linewidth]{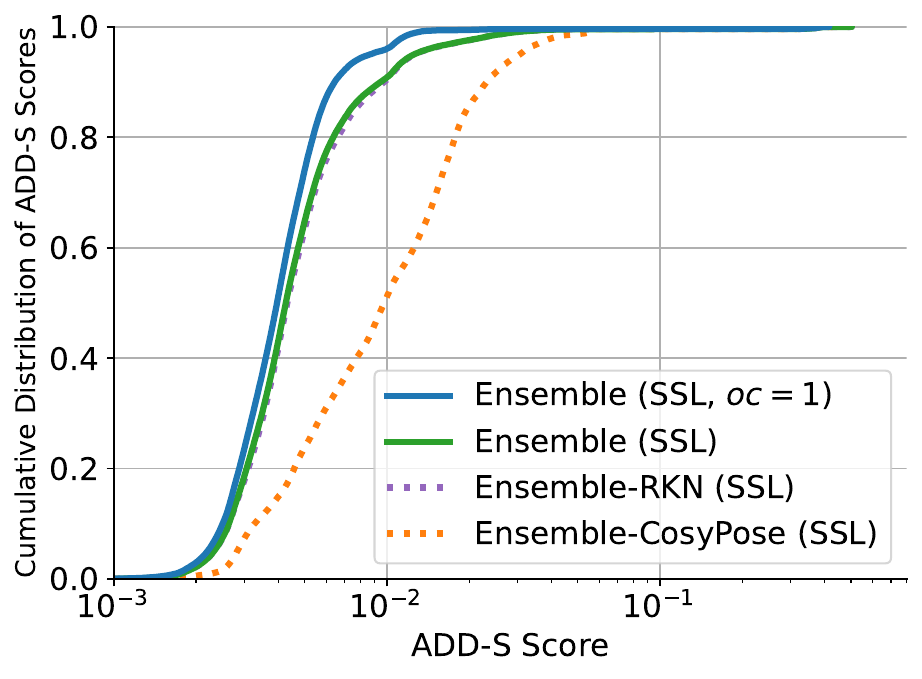}
				\caption{ }
		\label{fig:csy-cpo-contribution}
	\end{subfigure}
	\begin{subfigure}{0.25\textwidth}
		\begin{subfigure}{\textwidth}
			\includegraphics[trim=0 0 0 0,clip,width=\linewidth]{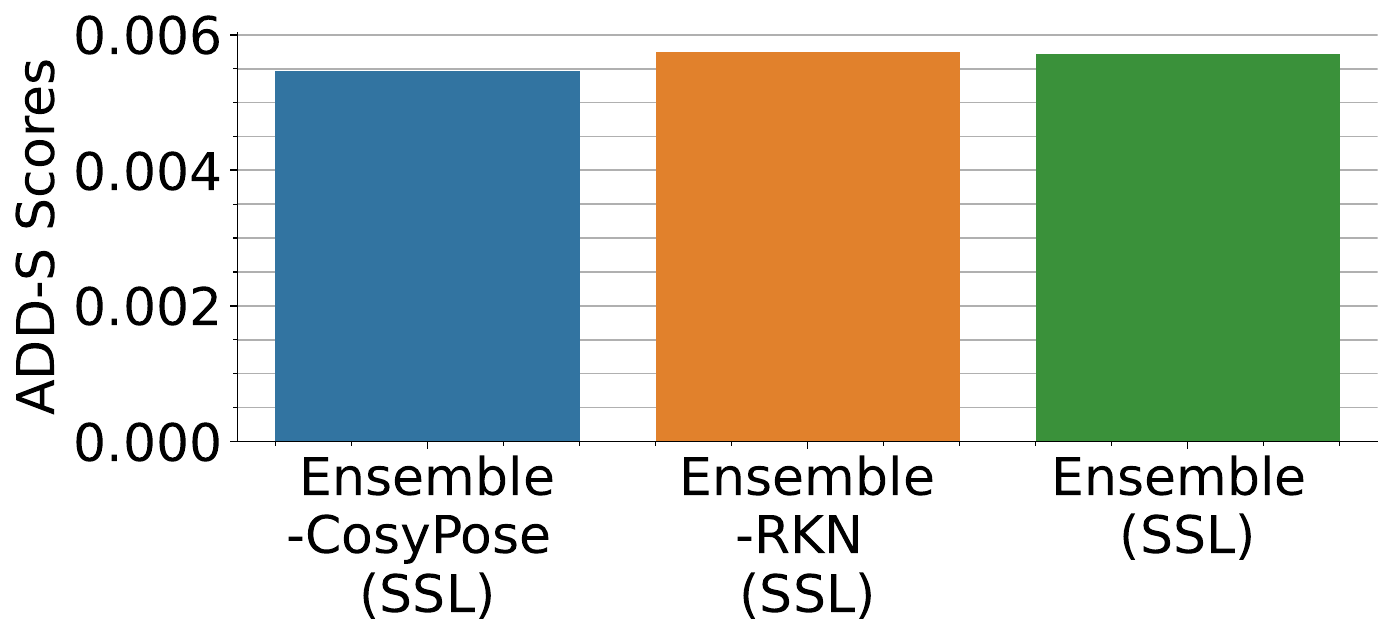}
					\caption{ }
			\label{fig:certifiable-adds}
		\end{subfigure}
		\begin{subfigure}{\textwidth}
			\includegraphics[trim=0 0 0 0,clip,width=\linewidth]{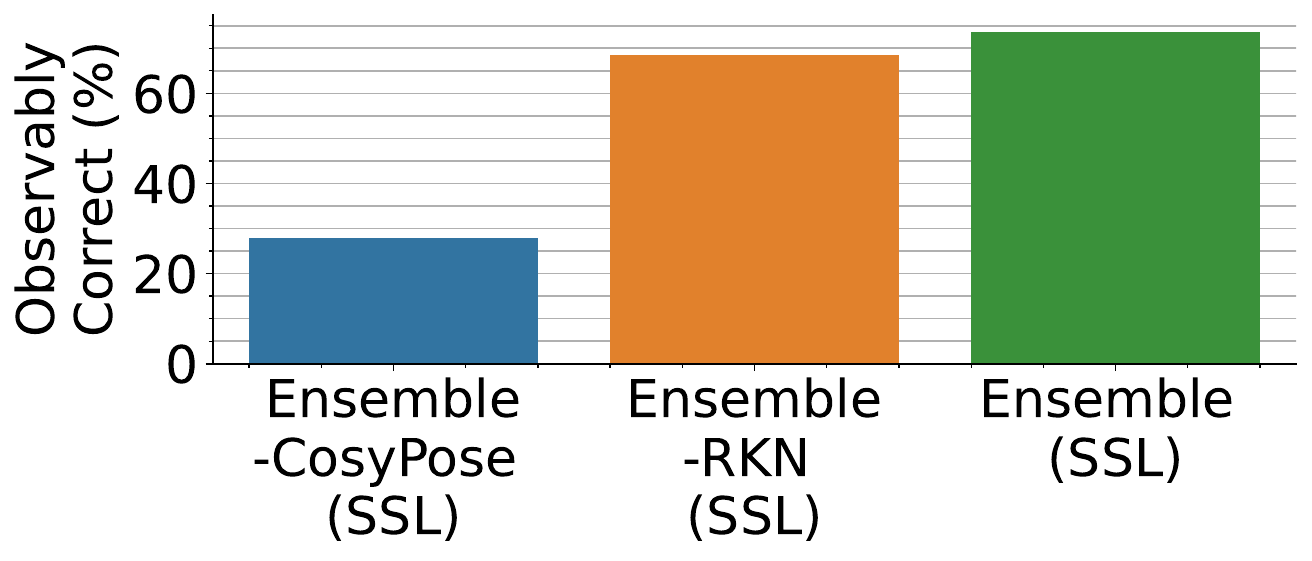}
					\caption{ }
			\label{fig:certifiable-inst}
		\end{subfigure}
	\end{subfigure}
	\caption{(a) Cumulative distribution of ADD-S scores, averaged across all objects in the \YCBV dataset, for the proposed \es architecture (\name) and other baselines. (b) Cumulative distribution of ADD-S scores, across all objects in the \YCBV dataset, for \name and its branches. (c) Average ADD-S score attained by the observably correct outputs of \name and its branches. (d) Percentage of observably correct instances for \name and its branches. \vspace{-4mm}}
	\label{fig:expt-ycb}
\end{figure*}

\myParagraph{Setup} We consider an \es architecture with two complementary pose estimators: \csy~\cite{Labbe20eccv-CosyPose}, a CNN-based, RGB-only pose estimator, and \rKeyPo, the point cloud-based pose estimator described in~\cref{sec:robust-cnp}. 
We take sim-trained \csy and \rKeyPo, 
and 
stack them 
in our \es architecture and self-train them on a real unannotated dataset, using the self-training method in \cref{sec:parallel}.
We label the resulting self-trained model ``\name (SSL)'' (as in self-supervised learning); we also report the accuracy achieved only by the outputs deemed observably correct: we denote the resulting results as ``\name (SSL, $\ocx=1$)''.
Moreover, we also evaluate the performance of each branch in the \es architecture. \nameCsy denotes the results corresponding to the corrected poses from the \csy branch. Similarly, \nameRky denotes the results corresponding to the corrected poses from the \rKeyPo branch.
 We compare these results against fully supervised versions of \csy and \rKeyPo, and against \selfSD~\cite{Wang22pami-OcclusionAwareSelfSupervised}, a state-of-the-art self-supervised method, with { weights and hyper-parameters provided by the authors}.

We train the supervised baselines on the \YCBV and T-LESS training set in the BOP dataset~\cite{Hodan20eccvw-BOPChallenge}; we use the given training and test split. For \es, the initial models are pre-trained on the synthetic data generated by a rendering engine, also provided in the BOP dataset. The models are then self-trained on the real test dataset. %
We use the stochastic gradient descent optimizer, with a learning rate of $2 \cdot 10^{-2}$ for \rKeyPo and $3 \cdot 10^{-4}$ for \csy for $20$ epochs over the standard test splits. Other hyper-parameters are reported in \arxivomit{the supplementary materials.}\arxivadd{Appendix~\ref{sec:hyperparam}.} 
For the supervised baselines, we use the suffix ``Synth.'' when they are trained on the synthetic data, or ``Real'' when they are trained on the real training data; for instance ``\csy~(Real)'' corresponds to the state-of-the-art approach from~\cite{Labbe20eccv-CosyPose}, which is trained in a supervised manner using the training data from \YCBV or \TLESS. 
For all approaches we use the 2D segmentation masks provided by the datasets (\cref{fig:pc-outliers}).
We measure performance using the ADD-S score~\cite{Wang19cvpr-DenseFusion6D}, computed on the test set and averaged across~all~objects. %

\myParagraph{Results and Insights} \cref{fig:expt-ycb}(a) shows the cumulative distribution of the ADD-S scores for each approach. The figure shows that the proposed \es outperforms by a small margin the fully supervised \rKeyPo (Real) and \csy (Real) approaches on the \YCBV test dataset. This shows that the \es architecture is able to reap the complementary benefits provided by the two models, and enhance them, without the need for external supervision. The figure also shows the performance of only the observably correct outputs produced by \name, \ie~\name (SSL, $\ocx=1$). We see that these exhibit a further performance boost, indicating that our certificates of observable correctness are indeed able to identify correct outputs. The figure also shows that \rKeyPo (Synth.) and \csy (Synth.), which are trained on synthetic data, perform poorly, indicating a large sim-to-real gap, and {remarking} the capability of the proposed \es to self-train starting from poor initial models.
{Finally, we observe that the proposed \es significantly outperforms \selfSD{ }--- a state-of-the-art, self-supervised pose estimation method~\cite{Wang22pami-OcclusionAwareSelfSupervised}.}

\Cref{fig:expt-ycb}(b) shows how each branch in the \es architecture contributes to its overall performance. We observe that \nameRky contributes the most. This is partly because \rKeyPo, in the \rKeyPo-branch, works directly on the RGB-D input, as opposed to \csy, which only relies on RGB information. However, a significant performance boost to \rKeyPo is provided by the robust corrector (\cf with~\cref{fig:expt-ycb}(a), where \rKeyPo (Real) does much worse than \csy (Real)).

While \cref{fig:expt-ycb}(a)-(b) showed that the observably correct outputs correspond to highly accurate pose estimates ---this is also true for each branch, see \cref{fig:expt-ycb}(c)--- not all outputs produced by \name are observably correct.~\Cref{fig:expt-ycb}(d) shows the percentage of observably correct outputs produced by \name, and each of its branches. 
We observe a gap in terms of the \% of observably correct instances produced. While the \rKeyPo-branch produces $67\%$ observably correct outputs, this number is  $28\%$ for the \csy branch.  
The overall \name architecture, as expected, gets the best of both models, and $70\%$ of its outputs are observably correct. 

\Cref{tab:expt_ycbv,tab:expt_tless} provide further insights into the compared techniques by breaking down the results by objects (\cref{fig:expt-ycb} instead averaged results across all objects).
The tables report the threshold ADD-S score
with a threshold equal to $5\%$ of the object diameter, and ADD-S (AUC) with a threshold equal to $10\%$ of the object diameter.
Note that this is a much stricter setup, compared to~\cite{Wang22pami-OcclusionAwareSelfSupervised}, which uses a much larger ADD-S threshold of $10$ cm.
\Cref{tab:expt_ycbv} shows the performance on six \YCBV objects, and~\Cref{tab:expt_tless} on six T-LESS objects; \arxivomit{extra results are given in the supplementary material.}\arxivadd{evaluation of all \YCBV and \TLESS objects is given in \Cref{tab:full_expt_ycbv,tab:full_expt_tless}.} 
We see that our conclusions, gleaned from~\cref{fig:expt-ycb}, still hold when parsing the object-specific performance. This reinforces that the proposed \es ensures very accurate pose estimates and \outperforms fully supervised baselines while not requiring real-world 3D annotations.
Qualitative results comparing our \name against the supervised baselines, \rKeyPo (Real) and \csy (Real), are given in~\cref{fig:front-page}.

\begin{table*}[t]

\centering
\caption{Evaluation of \name and baselines on the YCBV dataset.}
\label{tab:expt_ycbv}
\begin{tabular}{lrrrrrrrrrrrr}
	\toprule
	ADD-S~~~~ADD-S (AUC) &\multicolumn{2}{c}{Coffee Can}  &\multicolumn{2}{c}{Sugar Box}  &\multicolumn{2}{c}{Tuna Can}  &\multicolumn{2}{c}{Wood Block}  &\multicolumn{2}{c}{Scissors}  &\multicolumn{2}{c}{Large Clamp}  \\
	\midrule
    \rKeyPo (Real)   & 0.90 & 0.65 & 0.14 & 0.16 & 0.62 & 0.50 & 0.05 & 0.18 & 0.01 & 0.16 & 0.12 & 0.23 \\
    \csy (Real)      & 0.83 & 0.63 & 1.00 & 0.81 & 0.93 & 0.70 & 0.27 & 0.29 & 0.22 & 0.28 & 0.86 & 0.67 \\
    Self6D++ (SSL)   & 0.25 & 0.29 & 0.36 & 0.43 & 0.32 & 0.33 & 0.26 & 0.25 & 0.11	& 0.14 & 0.38 & 0.32 \\
    \name   (SSL)    & 1.00 & 0.77 & 0.99 & 0.79 & 1.00 & 0.76 & 0.98 & 0.69 & 0.96 & 0.73 & 0.97 & 0.78 \\
      \midrule
    \name (SSL, $\ocx=1$) & 1.00 & 0.78 & 1.00 & 0.83 & 1.00 & 0.79 & 1.00 & 0.74 & 1.00 & 0.77 & 0.97 & 0.78 \\
	\bottomrule
\end{tabular}
\end{table*}

\begin{table*}[t]

\centering
\caption{Evaluation of \name and baselines on the T-LESS dataset.}
\label{tab:expt_tless}
\begin{tabular}{lcccccccccccc}
	\toprule
	ADD-S~~~~ADD-S (AUC) &\multicolumn{2}{c}{{\tt obj_000004}}  &\multicolumn{2}{c}{{\tt obj_000010}}  &\multicolumn{2}{c}{{\tt obj_000013}}  &\multicolumn{2}{c}{{\tt obj_000024}}  &\multicolumn{2}{c}{{\tt obj_000026}}  &\multicolumn{2}{c}{{\tt obj_000030}}  \\
	\midrule
    \rKeyPo (Real)   & 0.19 & 0.32 & 0.62 & 0.52 & 0.46 & 0.48 & 0.31 & 0.39 & 0.40 & 0.47 & 0.79 & 0.58 \\
    \csy (Real)      & 0.41 & 0.38 & 0.84 & 0.63 & 0.51 & 0.42 & 0.62 & 0.50 & 0.83 & 0.64 & 0.93 & 0.71 \\
    \name  (SSL)     & 0.38 & 0.42 & 0.74 & 0.56 & 0.43 & 0.46 & 0.66 & 0.52 & 0.53 & 0.52 & 0.97 & 0.73 \\ \midrule
    \name (SSL, $\ocx=1$) & 0.85 & 0.65 & 0.99 & 0.73 & 0.79 & 0.59 & 0.98 & 0.69 & 1.00 & 0.72 & 1.00 & 0.75 \\
	\bottomrule
\end{tabular}
 	\vspace{-2mm}
\end{table*}

\begin{table}

\centering
\caption{{Effects of the robust corrector on closing the sim-to-real gap for both \csy and \rKeyPo.}}
\label{tab:corrector_sim2real}
\begin{tabular}{lccc}
	\toprule
	ADD-S & Synth.  & Synth. with Robust Corrector & Real \\
	\midrule 
	\csy  &  0.41 & 0.69  & 0.85 \\
	\rKeyPo &  0.17 & 0.36  & 0.36 \\
	\bottomrule
\end{tabular}

\end{table}

{
  \myParagraph{Insights: Impact of Corrector} The robust corrector in~\cref{sec:robustCorrector} was proposed to help bridge the sim-to-real gap.~\Cref{tab:corrector_sim2real} validates our proposal by showing the threshold ADD-S scores for the sim-trained models, \rKeyPo (Synth.) and \csy (Synth.), with and without the robust corrector, along with the fully supervised models on the real data, \ie \rKeyPo (Real) and \csy (Real).
  We see that the robust corrector provides the anticipated performance boost, and helps to bridge  the sim-to-real gap. Extra results with cumulative distributions of ADD-S scores are given in \arxivomit{the supplementary material.}\arxivadd{Appendix~\ref{sec:impactCorrector}.}   %
}

{
  \myParagraph{Insights: Progression of the \ES} 
  \Cref{fig:front-page} shows that the number of observably correct instances grows as the 
  \es progresses. \arxivomit{The supplementary material}\arxivadd{Appendix~\ref{sec:progressionES}} provides further results, showing the
   increase in the average number of observably correct instances for both branches of the ensemble, as well as breaking down the results in terms of 2D and 3D certificates.
}

\section{Conclusion}
\label{sec:conclusion}

We advance self-supervised learning for object pose estimation by proposing an \es architecture that simultaneously trains multiple models without manual 3D annotations, and leverages the complementarity of the models to further boost their performance. 
The \es architecture is enabled (i) by a differentiable robust corrector, which refines the pose estimates by each model, %
and (ii) by the definition of observable correctness certificates that identify correct pose estimates at test-time. 
The proposed \es performs on par or better compared to  state-of-the-art, fully supervised methods and largely outperforms competing self-supervised baselines.
As additional contributions, we introduce a robust centroid computation and robust pooling operation that 
empirically enhance the performance of point-cloud-based architectures for keypoint detection in the presence of outliers caused by a noisy 2D object segmentation. %

The results in this paper open several avenues of future work. First, 
it would be interesting to extend the proposed self-supervised approach to category-level perception where the object CAD model is unknown or has to be selected from a library of CAD models~\cite{Shi21rss-pace}. 
Second, it would be interesting to enhance the robust corrector to leverage 2D information (currently, it only uses the point cloud). Finally, the proposed robust centroid and robust pooling are expected to be useful for other point-cloud processing tasks, including point cloud segmentation and shape completion.

\arxivadd{
	\appendix
\arxivomit{ 
\section{The YCBV and TLESS Experiments}
\label{supp:sec:ycbv-tless-expt}

\subsection{Additional Results}
\label{sec:additionalResults}
\Cref{tab:expt_ycbv,tab:expt_tless} in the main paper analyzed performance of the pose estimation methods for selected objects in the \YCBV and \TLESS datasets, and only showed a subset of them due to space constraints. \Cref{tab:full_expt_ycbv,tab:full_expt_tless} show the same evaluations for all objects in the \YCBV and \TLESS datasets, respectively. 
The insights remain the same as noted in the paper. %
}

\subsection{Impact of the Robust Corrector}
\label{sec:impactCorrector}

The proposed robust corrector is designed to help bridge the sim-to-real gap.~\Cref{tab:corrector_sim2real} in the main paper validates our proposal by showing the threshold ADD-S scores for the sim-trained models, \rKeyPo (Synth.) and \csy (Synth.), with and without the robust corrector, along with the fully supervised models on the real data, \ie \rKeyPo (Real) and \csy (Real).
\Cref{fig:corrector_sim2real} plots the full distribution of the ADD-S scores for these models, further confirming that the corrector does indeed help bridge the sim-to-real gap.

\begin{figure}
	\centering 
	\begin{subfigure}{0.35\textwidth}
		\includegraphics[trim=0 0 0 0,clip,width=\linewidth]{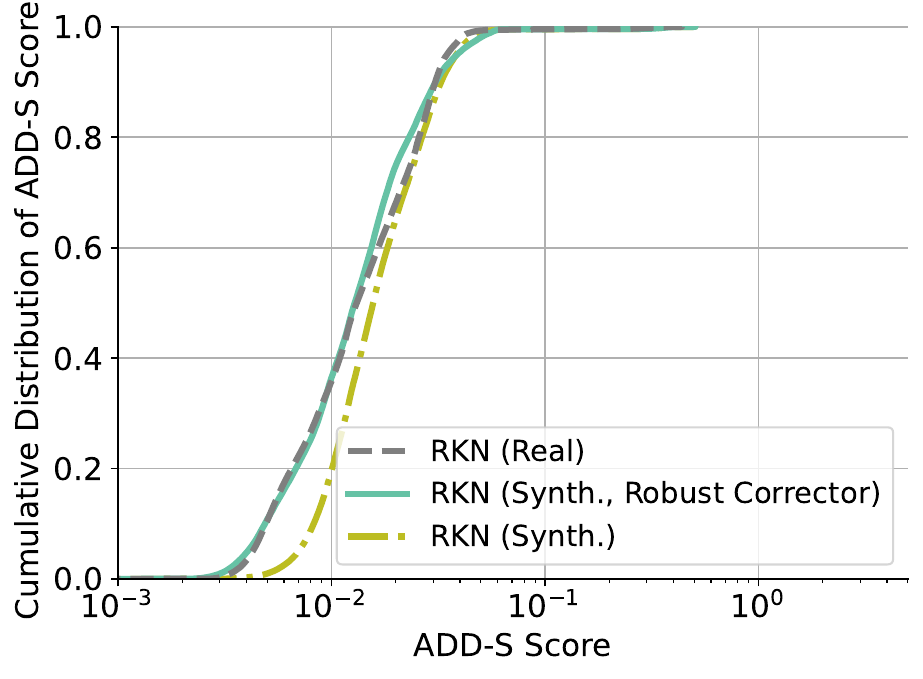}
		\caption{ } 
	\end{subfigure}
	\begin{subfigure}{0.35\textwidth}
		\includegraphics[trim=0 0 0 0,clip,width=\linewidth]{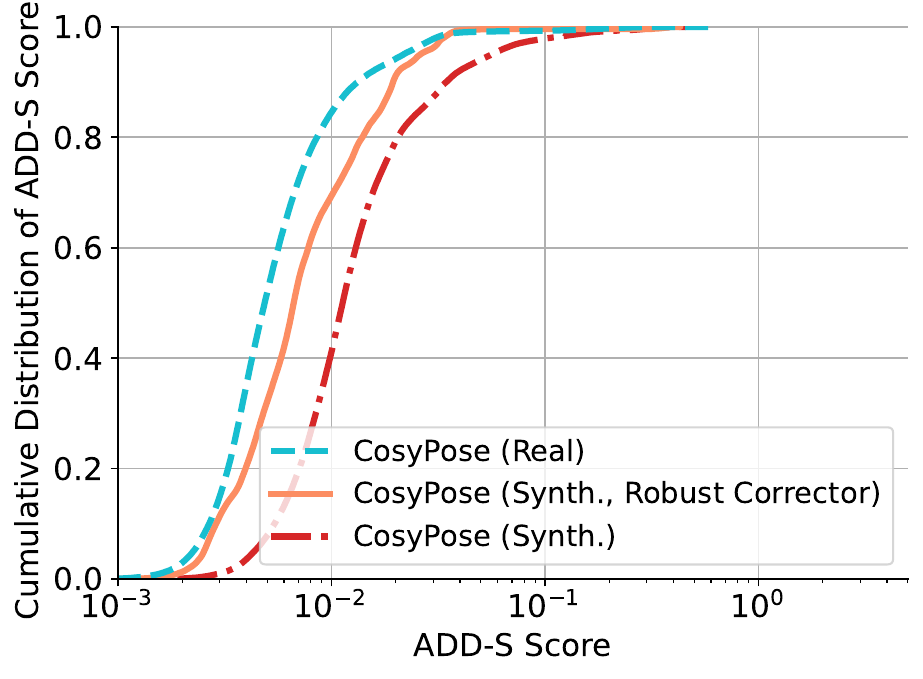}
	\caption{ } 
\end{subfigure}
\caption{Cumulative distribution of ADD-S scores, averaged across all objects in the \YCBV dataset. (a) The robust Corrector bridges the sim-to-real gap for \rKeyPo. (b) The robust Corrector partially bridges the sim-to-real gap for \csy.}
\label{fig:corrector_sim2real}
\end{figure}

\subsection{Progression of the \ES}
\label{sec:progressionES}
\Cref{fig:front-page} in the main paper showed that the number of observably correct instances grows as the 
\es progresses, confirming the effectiveness of the proposed self-training.
\Cref{fig:cert-train} provides further insights by showing the average increase in the percentage of 
instances with $(\ocTwoDx=1)$ and $(\ocThreeDx=1)$, produced by both the branches, after our \es. %
For \csy, we also consider a variant that implements a trainable coarse detector, which directly regresses object pose, while the \csy (Refine) implements a pre-trained coarse detector along with a trainable pose refinement model~\cite{Li18eccv-DeepIMDeep}. \csy (Refine), in fact, is what is proposed in~\cite{Labbe20eccv-CosyPose}
to achieve high accuracy.
We observe that the \rKeyPo achieves the highest increase: $23\%$ for $(\ocTwoDx=1)$ and $56\%$ for $(\ocThreeDx=1)$.
We also see an increase for the \csy branch. 
The \csy (Coarse) shows a higher improvement ($10\%$ and $24\%$) compared to \csy (Refine) ($1\%$ and $3\%$). %
Despite the modest increase in \csy (Refine), it still shows higher pose estimation accuracy after self-training, compared to \csy (Coarse). Therefore, in the baseline comparisons we only show \csy (Refine). 
\begin{figure}
	\centering 
	\includegraphics[trim=0 0 0 0,clip,width=\linewidth]{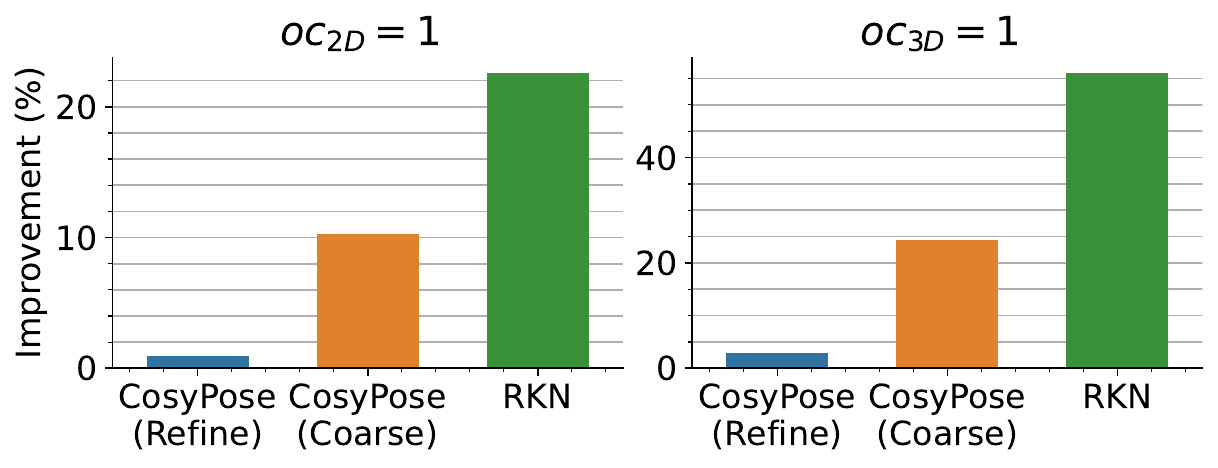}
	\caption{ {Average percentage increase in the number of instances with $(\ocThreeDx=1)$ and $(\ocTwoDx=1)$, after \es. Results are for the \rKeyPo branch, and two variations of the \csy branch, namely,  \csy (Coarse) and \csy (Refine).   %
	}}
	\vspace{-5mm}
	\label{fig:cert-train}
\end{figure}

\subsection{List of Hyper-Parameters}
\label{sec:hyperparam}
The proposed \es used stochastic gradient descent optimizer with a learning rate of $2 \cdot 10^{-2}$ for \rKeyPo and $3 \cdot 10^{-4}$ for \csy. The momentum and weight decay was set to $0.9$ and $1 \cdot 10^{-5}$, respectively, and a batch size of $20$ was used during the \es. 
The clamp threshold $\bar{c}$ in the robust corrector (eq.~\eqref{eq:corrector} in the main paper) and the loss function
(eq.~\eqref{eq:loss-self} in the main paper) was set to $10\%$ of the object diameter. 
The $\epsilon_\ocThreeDx$ and $\epsilon_\ocTwoDx$, used in the 3D and 2D certificates (eqs.~\eqref{eq:cert3d}-\eqref{eq:cert2d} in the main paper), were tuned manually for different objects. 
For \YCBV, $\epsilon_\ocThreeDx$ was set to $4\%$ of the object diameter, whereas $\epsilon_\ocTwoDx$ was chosen to be the best among: $50\%$, $60\%$, and $95\%$ of the object diameter. 
For \TLESS, $\epsilon_\ocTwoDx$ was set to $90\%$ of the object diameter, whereas $\epsilon_\ocThreeDx$ was chosen to be the best among: $4\%$, $6\%$, and $8\%$ of the object diameter. 
We had to be lenient in the choice of $\epsilon_\ocTwoDx$ because of the inaccuracies in the detected masks.
\Cref{tab:epsilon} lists them for all the \YCBV and \TLESS objects.

 \begin{table*}[t]

\centering
\caption{Parameters $\epsilon_\ocThreeDx$ and $\epsilon_\ocTwoDx$, as a $\%$ of the object diameter, used for the \YCBV and \TLESS objects.}
\label{tab:epsilon}
\begin{tabular}{|c|lrr|lrr|lrr|lrr|}
	\toprule
	\multirow{8}{*}{\rotatebox{90}{\hspace{-5mm}\YCBV} }  &
	object & $\epsilon_\ocThreeDx$  & $\epsilon_\ocTwoDx$  &   object & $\epsilon_\ocThreeDx$  & $\epsilon_\ocTwoDx$   &   object & $\epsilon_\ocThreeDx$  & $\epsilon_\ocTwoDx$   &   object & $\epsilon_\ocThreeDx$  & $\epsilon_\ocTwoDx$   \\
	\midrule
	&
	\texttt{obj_01} &  4\% & 95\% &
	\texttt{obj_02} &  4\% & 95\% &
	\texttt{obj_03} &  4\% & 95\% &
	\texttt{obj_04} &  4\% & 95\% \\
	&\texttt{obj_05} &  4\% & 60\% & 
	\texttt{obj_06} &  4\% & 95\% &
	\texttt{obj_07} &  4\% & 50\% &
	\texttt{obj_08} &  4\% & 95\% \\
	&\texttt{obj_09} &  4\% & 60\% & 
	\texttt{obj_10} &  4\% & 50\% &
	\texttt{obj_11} &  4\% & 95\% &
	\texttt{obj_12} &  4\% & 60\% \\
	&\texttt{obj_13} &  4\% & 95\% & 
	\texttt{obj_14} &  4\% & 95\% &
	\texttt{obj_15} &  4\% & 50\% &
	\texttt{obj_16} &  4\% & 95\% \\
	&\texttt{obj_17} &  4\% & 50\% & 
	\texttt{obj_18} &  4\% & 50\% &
	\texttt{obj_19} &  4\% & 50\% &
	\texttt{obj_20} &  4\% & 50\% \\
	&\texttt{obj_21} &  4\% & 60\% & 
	\texttt{--} & && 
	\texttt{--} & &&
	\texttt{--} & & \\
	\bottomrule
\end{tabular}

\vspace{0.3cm}
\begin{tabular}{|c|lrr|lrr|lrr|lrr|}
	\toprule
	\multirow{8}{*}{\rotatebox{90}{\hspace{-5mm}\TLESS} } 
	&
	object & $\epsilon_\ocThreeDx$  & $\epsilon_\ocTwoDx$  &   object & $\epsilon_\ocThreeDx$  & $\epsilon_\ocTwoDx$   &   object & $\epsilon_\ocThreeDx$  & $\epsilon_\ocTwoDx$   &   object & $\epsilon_\ocThreeDx$  & $\epsilon_\ocTwoDx$   \\
	\midrule
	& \texttt{obj_01} &  4\% & 90\% &
	\texttt{obj_02} &  4\% & 90\% &
	\texttt{obj_03} &  4\% & 90\% &
	\texttt{obj_04} &  4\% & 90\% \\
	& \texttt{obj_05} &  6\% & 90\% & 
	\texttt{obj_06} &  6\% & 90\% &
	\texttt{obj_07} &  6\% & 90\% &
	\texttt{obj_08} &  6\% & 90\% \\
	& \texttt{obj_09} &  6\% & 90\% & 
	\texttt{obj_10} &  6\% & 90\% &
	\texttt{obj_11} &  6\% & 90\% &
	\texttt{obj_12} &  6\% & 90\% \\
	&\texttt{obj_13} &  4\% & 90\% & 
	\texttt{obj_14} &  4\% & 90\% &
	\texttt{obj_15} &  8\% & 90\% &
	\texttt{obj_16} &  8\% & 90\% \\
	&\texttt{obj_17} &  4\% & 90\% & 
	\texttt{obj_18} &  4\% & 90\% &
	\texttt{obj_19} &  6\% & 90\% &
	\texttt{obj_20} &  6\% & 90\% \\
	&\texttt{obj_21} &  6\% & 90\% & 
	\texttt{obj_22} &  6\% & 90\% &
	\texttt{obj_23} &  6\% & 90\% &
	\texttt{obj_24} &  4\% & 90\% \\
	&\texttt{obj_25} &  4\% & 90\% & 
	\texttt{obj_26} &  4\% & 90\% &
	\texttt{obj_27} &  6\% & 90\% &
	\texttt{obj_28} &  6\% & 90\% \\
	&\texttt{obj_29} &  4\% & 90\% & 
	\texttt{obj_30} &  4\% & 90\% & 
	\texttt{--} &  & &
	\texttt{--} & & \\
	\bottomrule
\end{tabular} \end{table*}

\begin{table*}

\centering
\caption{Evaluation of \name and baselines on 21 YCBV objects.}
\label{tab:full_expt_ycbv}

\begin{tabular}{lcccccccccccc}
	\toprule
	ADD-S~~~~ADD-S (AUC) &\multicolumn{2}{c}{{Coffee Can}}  &\multicolumn{2}{c}{{Cracker Box}}  &\multicolumn{2}{c}{{Sugar Box}}  &\multicolumn{2}{c}{{Soup Can}}  &\multicolumn{2}{c}{{Mustard Bottle}}  &\multicolumn{2}{c}{{Tuna Can}}  \\
	\midrule
	\rKeyPo (Real)   & 0.90	& 0.65 & 0.01 & 0.06 & 0.14	& 0.16 & 0.64 &	0.51 & 0.64	& 0.47 & 0.62 & 0.50 \\
	\csy (Real)      & 0.83	& 0.63 & \bf 0.98 &	\bf 0.75 & \bf 1.00	& \bf 0.81 & 0.85 & 0.65 & 1.00	& 0.78 & 0.93 &	0.70 \\
	Self6D++         & 0.25	& 0.29 & 0.77 &	0.63 & 0.36	& 0.43 & 0.09 & 0.17 & 0.80	& 0.67 & 0.32 & 0.33 \\
	\name  (SSL)     & \bf 1.00	& \bf 0.77 & 0.68 &	0.55& 0.99 &	0.79&	\bf 0.96&	\bf 0.74&\bf	1.00&	\bf 0.78 & \bf 1.00 & \bf 0.76 \\ \midrule
	\name (SSL, $\ocx=1$) & 1.00 & 0.78&	0.96&	0.75&	1.00&	0.83&	0.98&	0.76&	1.00&	0.78&	1.00 & 0.79 \\
	
	\midrule[2pt]
	
	ADD-S~~~~ADD-S (AUC) &\multicolumn{2}{c}{{Pudding Box}}  &\multicolumn{2}{c}{{Gelatin Box}}  &\multicolumn{2}{c}{{Meat Can}}  &\multicolumn{2}{c}{{Banana}}  &\multicolumn{2}{c}{{Pitcher}}  &\multicolumn{2}{c}{{Bleach}}  \\
	\midrule
	\rKeyPo (Real)   & 0.01 &	0.18&	0.00&	0.15&	0.40&	0.35&	0.18&	0.23&	0.70&	0.55&	0.39&	0.35 \\
	\csy (Real)      & 0.91	&0.79&	0.88	&0.70&	0.61&	0.51&	0.94&	0.77&	\bf1.00&	\bf0.84&	0.82&	0.67 \\
	Self6D++         & 0.63	&0.58&	0.05&	0.16&	0.30&	0.30&	0.90&	0.68&	1.00&	0.77&	0.77&	0.62 \\
	\name  (SSL)     & \bf1.00	&\bf0.80&	\bf1.00&	\bf0.85&	\bf0.71&\bf	0.59&	\bf1.00&	\bf0.79&	0.81&	0.66&	\bf1.00&	\bf0.76 \\ \midrule
	\name (SSL, $\ocx=1$) & 1.00&	0.80&	1.00&	0.85&	0.80&	0.65&	1.00&	0.79&	1.00&	0.81&	1.00&	0.76 \\
	
	\midrule[2pt]
	
	ADD-S~~~~ADD-S (AUC) &\multicolumn{2}{c}{{Bowl}}  &\multicolumn{2}{c}{{Mug}}  &\multicolumn{2}{c}{{Power Drill}}  &\multicolumn{2}{c}{{Wood Block}}  &\multicolumn{2}{c}{{Scissors}}  &\multicolumn{2}{c}{{Marker}}  \\
	\midrule
	\rKeyPo (Real)   & 0.17&	0.14&	0.54&	0.52&	0.27&	0.28&	0.05&	0.18&	0.01&	0.16&	0.05&	0.06 \\
	\csy (Real)      & 0.48&	0.39&	0.81&	0.66&	0.98&	0.79&	0.27&	0.29&	0.22&	0.28&	0.50&	0.44 \\
	Self6D++         & 0.16&	0.24&	0.03&	0.17&	0.77&	0.62&	0.26&	0.25&	0.11&	0.14&	0.18&	0.17 \\
	\name  (SSL)     & \bf0.98&	\bf0.82&	\bf0.95&	\bf0.73&\bf	1.00&\bf	0.84&\bf	0.98&\bf	0.69&\bf	0.96&\bf	0.73&\bf	0.97&\bf	0.75 \\ \midrule
	\name (SSL, $\ocx=1$) & 0.98&	0.82&	1.00&	0.82&	1.00&	0.84&	1.00&	0.74&	1.00&	0.77&	0.99&	0.80 \\
	
	\midrule[2pt]
	
	ADD-S~~~~ADD-S (AUC) &\multicolumn{2}{c}{{ Large Clamp}}  &\multicolumn{2}{c}{{Extra Large Clamp}}  &\multicolumn{2}{c}{{Foam Brick}}  &\multicolumn{2}{c}{{\tt --}}  &\multicolumn{2}{c}{{\tt --}}  &\multicolumn{2}{c}{{\tt --}}  \\
	\midrule
	\rKeyPo (Real)   & 0.12&	0.23&	0.05&	0.19&	0.00&	0.20 &&&&&& \\
	\csy (Real)      & 0.86&	0.67&	\bf1.00&\bf	0.81&	0.54&	0.46 &&&&&& \\
	Self6D++         & 0.38&	0.32&	0.82&	0.63&	0.19&	0.24 &&&&&& \\
	\name  (SSL)     &\bf 0.97&\bf	0.78&	0.91&	0.73&	\bf0.59&\bf	0.51 &&&&&& \\ \midrule
	\name (SSL, $\ocx=1$) & 0.97&	0.78&	0.96&	0.77&	0.49&	0.43 &&&&&&  \\
	
	\bottomrule[2pt]
\end{tabular}
 \end{table*}

\begin{table*}

\centering
\caption{Evaluation of \name and baselines on 30 \TLESS objects.}
\label{tab:full_expt_tless}
\begin{tabular}{lcccccccccccc}
	\toprule
	ADD-S~~~~ADD-S (AUC) &\multicolumn{2}{c}{{\tt obj_000001}}  &\multicolumn{2}{c}{{\tt obj_000002}}  &\multicolumn{2}{c}{{\tt obj_000003}}  &\multicolumn{2}{c}{{\tt obj_000004}}  &\multicolumn{2}{c}{{\tt obj_000005}}  &\multicolumn{2}{c}{{\tt obj_000006}}  \\
	\midrule
    \rKeyPo (Real)   & 0.71&	0.50&	0.76&	0.53&	0.90&	0.61&	0.19&	0.32&	0.29&	0.40&	0.33&	0.43 \\
    \csy (Real)      & \bf0.60	& \bf0.45&	0.51&	0.40&	\bf0.84&	\bf0.62&	\bf0.41&	0.38&	\bf0.89&	\bf0.68&	\bf0.94&\bf	0.70 \\
    \name  (SSL)     & 0.51&	0.44&	\bf0.68&\bf	0.48&	0.69&	0.51&	0.38&	\bf0.42&	0.36&	0.43&	0.21&	0.36\\ \midrule
    \name (SSL, $\ocx=1$) & 0.33&	0.33&	0.40&	0.33&	0.32&	0.44&	0.85&	0.65&	0.71&	0.49&	0.25&	0.40 \\
	
	\midrule[2pt]
	
	ADD-S~~~~ADD-S (AUC) &\multicolumn{2}{c}{{\tt obj_000007}}  &\multicolumn{2}{c}{{\tt obj_000008}}  &\multicolumn{2}{c}{{\tt obj_000009}}  &\multicolumn{2}{c}{{\tt obj_000010}}  &\multicolumn{2}{c}{{\tt obj_000011}}  &\multicolumn{2}{c}{{\tt obj_000012}}  \\
	\midrule
	\rKeyPo (Real)   & 0.28&	0.40&	0.31&	0.40&	0.33&	0.41&	0.62&	0.52&	0.68&	0.53&	0.74&	0.53 \\
	\csy (Real)      & \bf0.94& \bf	0.72&	\bf0.90&	\bf0.66&	0.90&	0.68&	0.84&	0.63&	\bf0.77&	\bf0.57&	\bf0.84&	\bf0.63 \\
	\name  (SSL)     & 0.25&	0.35&	0.23&	0.33&	\bf0.74&	\bf0.51&	\bf0.90&	\bf0.62&	0.70&	0.53&	0.72&	0.53 \\ \midrule
	\name (SSL, $\ocx=1$) & 0.25&	0.36&	0.25&	0.35&	0.76&	0.52&	0.93&	0.65&	0.72&	0.55&	0.77&	0.55 \\
	
	\midrule[2pt]
	
	ADD-S~~~~ADD-S (AUC) &\multicolumn{2}{c}{{\tt obj_000013}}  &\multicolumn{2}{c}{{\tt obj_000014}}  &\multicolumn{2}{c}{{\tt obj_000015}}  &\multicolumn{2}{c}{{\tt obj_000016}}  &\multicolumn{2}{c}{{\tt obj_000017}}  &\multicolumn{2}{c}{{\tt obj_000018}}  \\
	\midrule
	\rKeyPo (Real)   & 0.46	&0.48&	0.16&	0.26&	0.53&	0.50&	0.75&	0.55&	0.85&	0.62&	0.72&	0.55 \\
	\csy (Real)      & 0.51	&0.42&	0.71&	0.54&	\bf0.79&	\bf0.58&	\bf0.89&	\bf0.64&\bf	0.94&	\bf0.72&\bf	0.93&\bf	0.76 \\
	\name  (SSL)     & \bf0.74	&\bf0.55	&\bf0.74&\bf	0.54&	0.65&	0.54&	0.74&	0.53&	0.81&	0.59&	0.90&	0.64 \\ \midrule
	\name (SSL, $\ocx=1$) & 0.82&	0.59&	0.80&	0.58&	0.68&	0.55&	0.77&	0.55&	0.95&	0.66&	0.98&	0.70 \\
	
	\midrule[2pt]
	
	ADD-S~~~~ADD-S (AUC) &\multicolumn{2}{c}{{\tt obj_000019}}  &\multicolumn{2}{c}{{\tt obj_000020}}  &\multicolumn{2}{c}{{\tt obj_000021}}  &\multicolumn{2}{c}{{\tt obj_000022}}  &\multicolumn{2}{c}{{\tt obj_000023}}  &\multicolumn{2}{c}{{\tt obj_000024}}  \\
	\midrule
	\rKeyPo (Real)   & 0.54&	0.48&	0.11	&0.32&	0.32&	0.40&	0.31&	0.43&	0.32&	0.35&	0.31&	0.39 \\
	\csy (Real)      & \bf0.79	&\bf0.59&	\bf0.73&	\bf0.55&	\bf0.75&	\bf0.61&	\bf0.70&	\bf0.55&	\bf0.87&	\bf0.67&	0.62&	0.50 \\
	\name  (SSL)     & 0.40	&0.46&	0.13&	0.30&	0.41&	0.41&	0.52&	0.48&	0.66&	0.52&	\bf0.66&	\bf0.52 \\ \midrule
	\name (SSL, $\ocx=1$) & 0.40&	0.44&	0.26&	0.33&	0.38&	0.39&	0.61&	0.52&	0.75&	0.58&	0.98&	0.69 \\
	
	\midrule[2pt]
	
	ADD-S~~~~ADD-S (AUC) &\multicolumn{2}{c}{{\tt obj_000025}}  &\multicolumn{2}{c}{{\tt obj_000026}}  &\multicolumn{2}{c}{{\tt obj_000027}}  &\multicolumn{2}{c}{{\tt obj_000028}}  &\multicolumn{2}{c}{{\tt obj_000029}}  &\multicolumn{2}{c}{{\tt obj_000030}}  \\
	\midrule
	\rKeyPo (Real)   & 0.40&	0.46&	0.40&	0.47&	0.32&	0.40&	0.42&	0.42&	0.14&	0.34&	0.79&	0.58 \\
	\csy (Real)      &  \bf0.86&	\bf0.65	&\bf0.83&	\bf0.64&	\bf0.91&	\bf0.69&	\bf0.85&	\bf0.69&	\bf0.94&	\bf0.72&	0.93&	0.71 \\
	\name  (SSL)     & 0.63	&0.55&	0.82&	0.61&	0.79&	0.57&	0.65&	0.46&	0.58&	0.49&	\bf0.97&	\bf0.73 \\ \midrule
	\name (SSL, $\ocx=1$) & 0.82&	0.60&	0.90&	0.64&	0.85&	0.60&	0.72&	0.51&	0.76&	0.59&	1.00&	0.75 \\
	
	\bottomrule[2pt]
\end{tabular} \end{table*}

\subsection{Additional Visualization}
\label{sec:addVis}

\Cref{fig:vis-collage} shows some examples of pose estimates resulting from the proposed \name and other baseline methods.

\begin{figure*}
	\centering
	\includegraphics[trim=0 0 0 0,clip,width=0.8\linewidth]{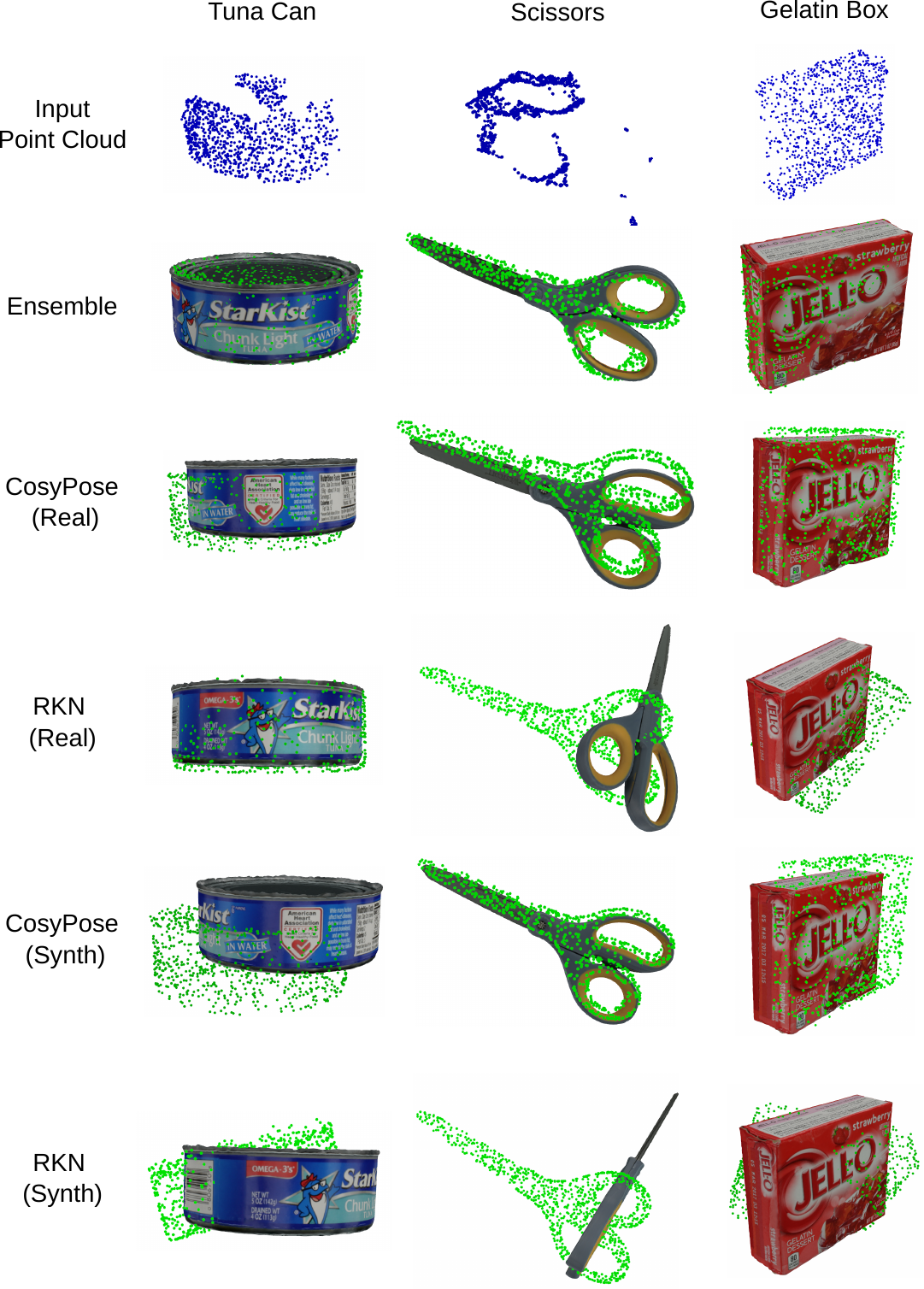}
	\caption{Sample estimates of the proposed \es (label: \Ensemble) and various baseline methods on three objects from the \YCBV dataset. The first row displays input point clouds generated from depth maps using segmentation masks. Green dots represent model point clouds transformed with ground-truth transformations, and the CAD models are transformed using the pose estimates.}
	\label{fig:vis-collage}
\end{figure*}
 }

\bibliographystyle{IEEEtran}

\end{document}